%% file: arxiv.tex
\documentclass[twocolumn, twoside]{IEEEtran}
\usepackage{cite}
\usepackage{amsmath,amssymb,amsfonts}
\usepackage{textcomp}
\usepackage{hyperref}
\usepackage{soul}
\usepackage{multirow}
\usepackage{subcaption}
\usepackage{url}
\usepackage{hyperref}
\usepackage{caption}
\usepackage{algorithmic}
\usepackage{graphicx}
\usepackage{textcomp}
\usepackage{mwe}
\usepackage{adjustbox}
\usepackage{rotating}
\usepackage{graphicx}
\usepackage{booktabs}

\newcommand{\vecg}{\mathbf{g}}
\newcommand{\vecx}{\mathbf{x}}
\newcommand{\vech}{\mathbf{h}}
\newcommand{\vecmu}{\boldsymbol{\mu}}
\newcommand{\best}[1]{\textbf{#1}}
\newcommand{\second}[1]{\underline{#1}}
\newcommand{\headeretal}{et al.}

\usepackage{bm}
\makeatletter
\AtBeginDocument{\DeclareMathVersion{bold}
\SetSymbolFont{operators}{bold}{T1}{times}{b}{n}
\SetMathAlphabet{\mathrm}{bold}{T1}{times}{b}{n}
\SetMathAlphabet{\mathit}{bold}{T1}{times}{b}{it}
\SetMathAlphabet{\mathbf}{bold}{T1}{times}{b}{n}
\SetMathAlphabet{\mathtt}{bold}{OT1}{pcr}{b}{n}
\SetSymbolFont{symbols}{bold}{OMS}{cmsy}{b}{n}
\renewcommand\boldmath{\@nomath\boldmath\mathversion{bold}}}
\makeatother

\def\BibTeX{{\rm B\kern-.05em{\sc i\kern-.025em b}\kern-.08em
    T\kern-.1667em\lower.7ex\hbox{E}\kern-.125emX}}

\begin{document}

\title{Looking around you: external information enhances representations for event sequences}


\author{
\thanks{The work was supported by the grant for research centers in the field of AI provided by the Ministry of Economic Development of the Russian Federation in accordance with the agreement 000000C313925P4F0002 and the agreement with Skoltech №139-10-2025-033.
}
\IEEEauthorblockN{Petr Sokerin$^{1}$, Maria Kovaleva$^{1}$, Ekaterina Boyarina$^{1}$, Pavel Tikhomirov$^{1}$,  Denis Vorobiyov$^{1,2}$, Alexey Zaytsev$^{1}$}

\IEEEauthorblockA{$^{1}$ LARSS Laboratory, AI Center, Skoltech, Bolshoy Boulevard, 30, bld. 1, Moscow, 121205, Russia}

\IEEEauthorblockA{$^{2}$ MIPT, IITP RAS, Institutsky Lane, 9, Dolgoprudny, Moscow Region, 141701, Russia}
}


\markboth
{Sokerin \headeretal: Looking around you: external information enhances representations for event sequences}
{Sokerin \headeretal: Looking around you: external information enhances representations for event sequences}





\maketitle

\begin{IEEEkeywords}
External information, external context aggregation, embeddings of sequences, sequential data, self-supervised learning.  
\end{IEEEkeywords}

\begin{abstract}
\input{chapters_sigmod/abstract}

\end{abstract}

\input{chapters_sigmod/01_intro}

\input{chapters_sigmod/02_literature}

\input{chapters_sigmod/03_methods}

\input{chapters_sigmod/04_numerical_experiments}


\input{chapters_sigmod/05_conclusion}

\bibliographystyle{ieeetr}
\bibliography{sample-base}

\clearpage
\appendix

\input{chapters_sigmod/99_appendix}
\newpage
\begin{IEEEbiography}[{\includegraphics[width=1in,height=1.25in,clip,keepaspectratio]{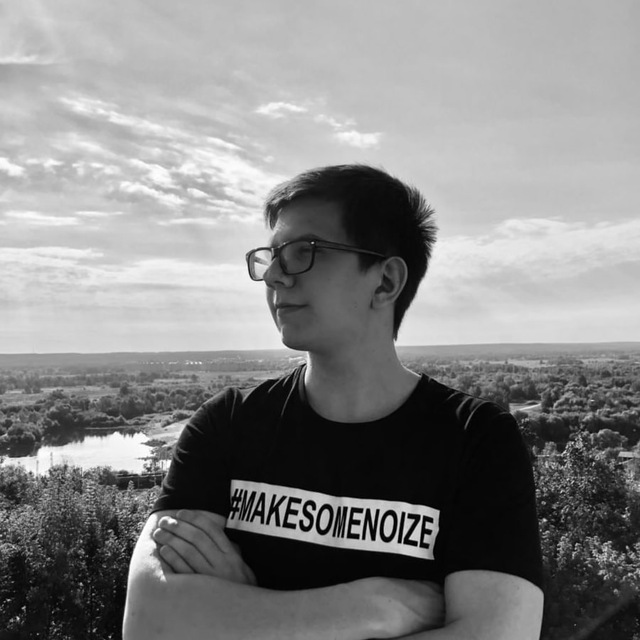}}]
{Petr Sokerin} 
got a bachelor's degree in PREU in 2021. 
He graduated from Skoltech in 2023 with a thesis on concealed adversarial attacks for time-series data. 
Now Petr works as a research engineer at Skoltech. 
Mr. Sokerin focuses his research on the development of new methods for sequential data and recommendation systems.
\end{IEEEbiography}

\begin{IEEEbiography}[{\includegraphics[width=1in,height=1.25in,clip,keepaspectratio]{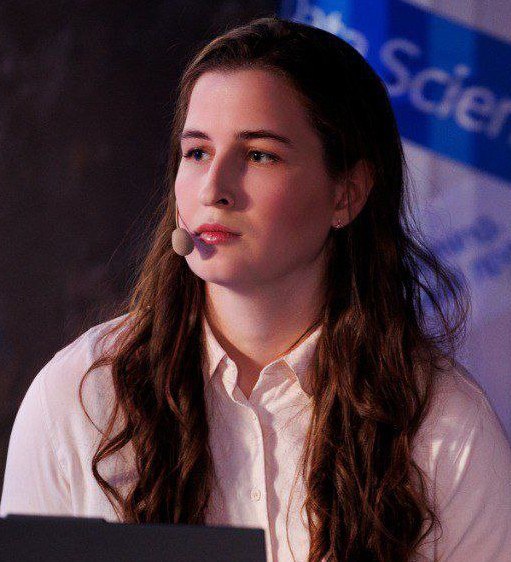}}]{Maria Kovaleva} received B.S. and M.S. degrees (Hons.) in applied mathematics and physics from the MIPT, in 2022 and 2024, respectively, and an M.S. degree (Hons.) in data science from the Skoltech, in 2024. Since 2024, she has been a Research Engineer at Kandinsky Lab, working in the field of generative computer vision and video generation. The work was conducted by the author while studying at Skoltech.
\end{IEEEbiography}

\begin{IEEEbiography}[{\includegraphics[width=1in,height=1.25in,clip,keepaspectratio]{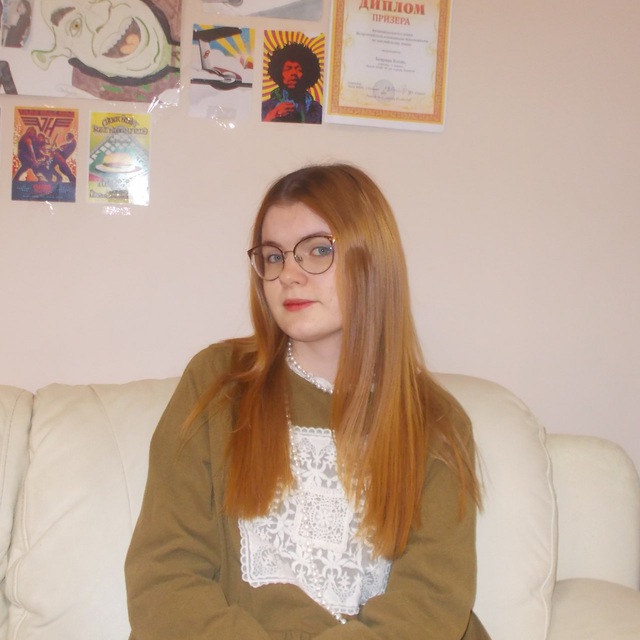}}]{Ekaterina Boyarina} is currently pursuing a bachelor’s degree in applied mathematics and physics at Moscow Institute of Physics and Technology (MIPT). She is a Research Intern at Skoltech, where she focuses on uncertainty estimation and neural networks for sequential data.
\end{IEEEbiography}

\begin{IEEEbiography}[{\includegraphics[width=1in,height=1.25in,clip,keepaspectratio]{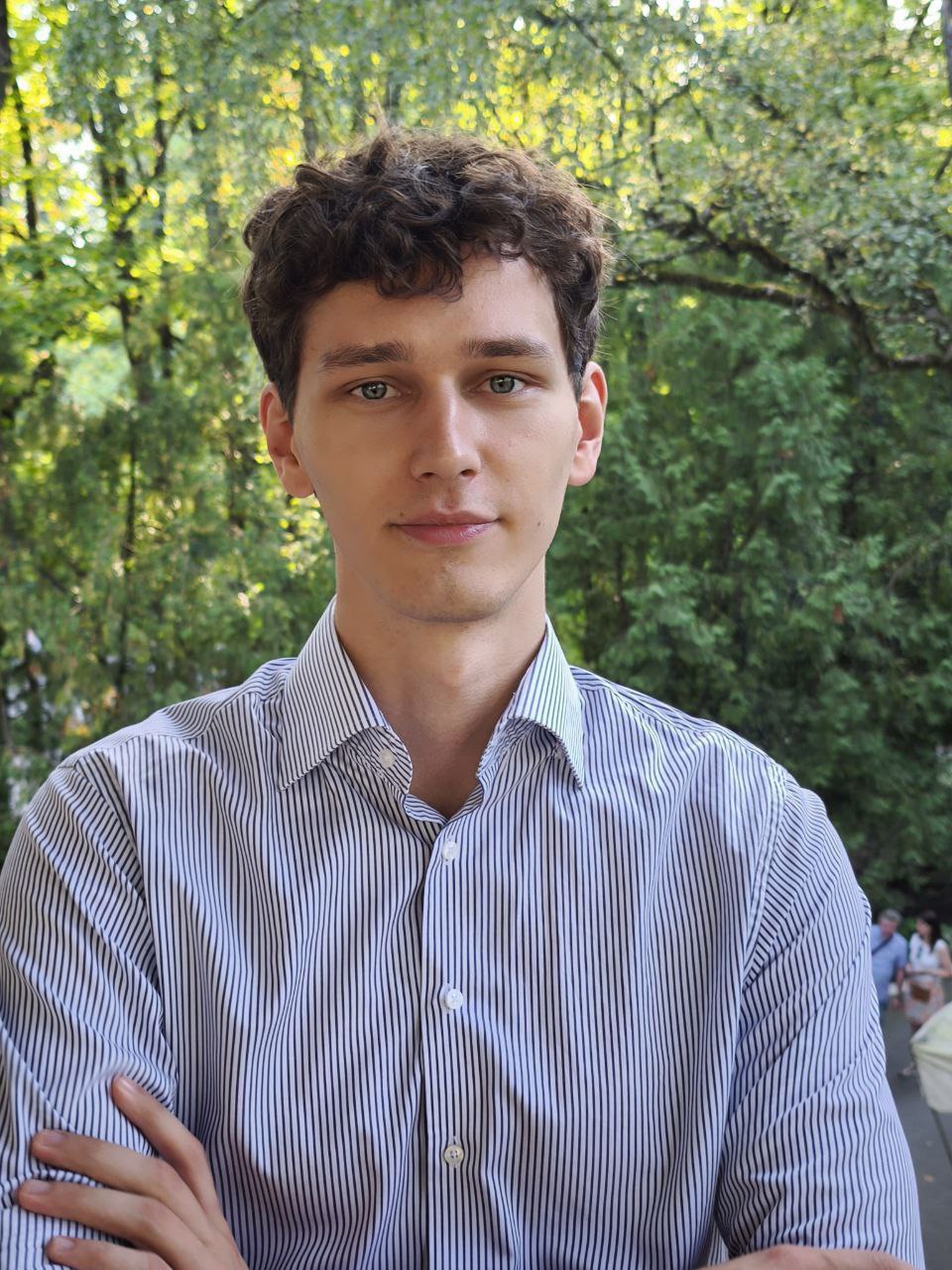}}]{Pavel Tikhomirov} received the bachelor's degree from NRNU MEPhI in 2023, and the degree from Skoltech in 2023 with a thesis on the application of LLMs in group theory. Currently, he is a Machine Learning engineer at Yandex. He focuses on the development of methods for search and recommendation systems.
\end{IEEEbiography}

\begin{IEEEbiography}[{\includegraphics[width=1in,height=1.25in,clip,keepaspectratio]{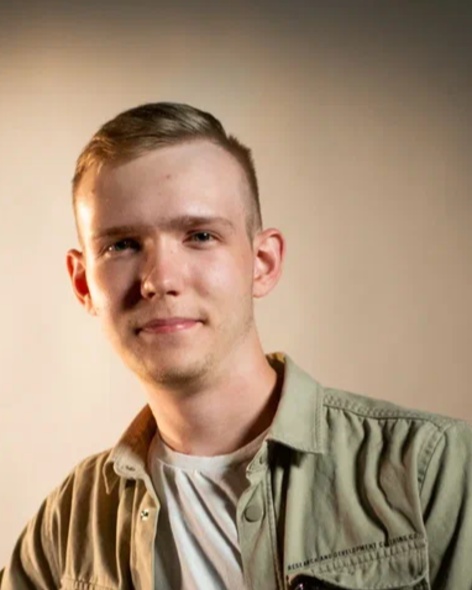}}]{DENIS VOROBEV} is currently pursuing the bachelor’s degree in applied mathematics and computer science at the Moscow Institute of Physics and Technology (MIPT). He is a Research Intern with the AI Center, Skoltech, where he focuses on adversarial attacks and neural networks for sequential data.
\end{IEEEbiography}

\begin{IEEEbiography}[{\includegraphics[width=1in,height=1.25in,clip,keepaspectratio]{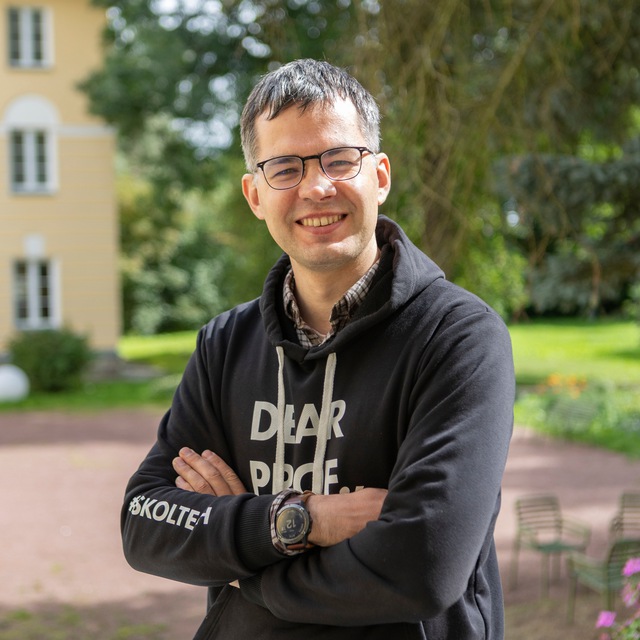}}]{Alexey Zaytsev} graduated from MIPT in 2012. He completed a Ph.D. in Math at IITP RAS in 2017 with a theoretical investigation of Gaussian process regression. Now Alexey works as an associate professor at Skoltech. Dr. Zaytsev focuses his research on the development of new methods for sequential data and uncertainty estimation.
\end{IEEEbiography}


\end{document}

%% file: chapters_sigmod/abstract.tex
Representation learning produces models in different domains, such as store purchases, client transactions, and general people's behavior. 
However, such models for event sequences usually process each sequence in isolation, ignoring context from those that co-occur in time.
This limitation is particularly problematic in domains with fast-evolving conditions, like finance and e-commerce, or when certain sequences lack recent events.

We develop a method that aggregates information from multiple user representations, augmenting a specific user's representation in a setting with multiple co-occurring event sequences, achieving better quality than processing each sequence independently.
Our study considers diverse aggregation approaches, ranging from simple pooling techniques to Learnable attention aggregation, that can highlight more complex information flow among other users.
The proposed methods operate on top of an existing encoder and support its efficient fine-tuning. 
Across nine diverse event sequence datasets (finance, e-commerce, entertainment, etc.) and downstream tasks, Learnable attention improves metric scores, both with and without fine-tuning, while mean pooling yields a smaller but still significant gain.

%% file: chapters_sigmod/01_intro.tex
\section{Introduction}

Humans and industrial processes~\cite{yan2019recent} naturally generate data of various forms that can be described as event sequences.
Daily financial transactions~\cite{babaev2019rnn}, purchases, and movie views, among others, appear as a rich source of information for data-based decision making~\cite{fan2014challenges}. 
Their inherent non-uniformity in time, heterogeneous pool of event types, and intricate between-events interactions motivate the development of continuous-time methods tailored to modeling event sequences~\cite{shchur2021neural}.

\begin{figure}[!t]
     \centering
     \includegraphics[width=0.9\columnwidth]{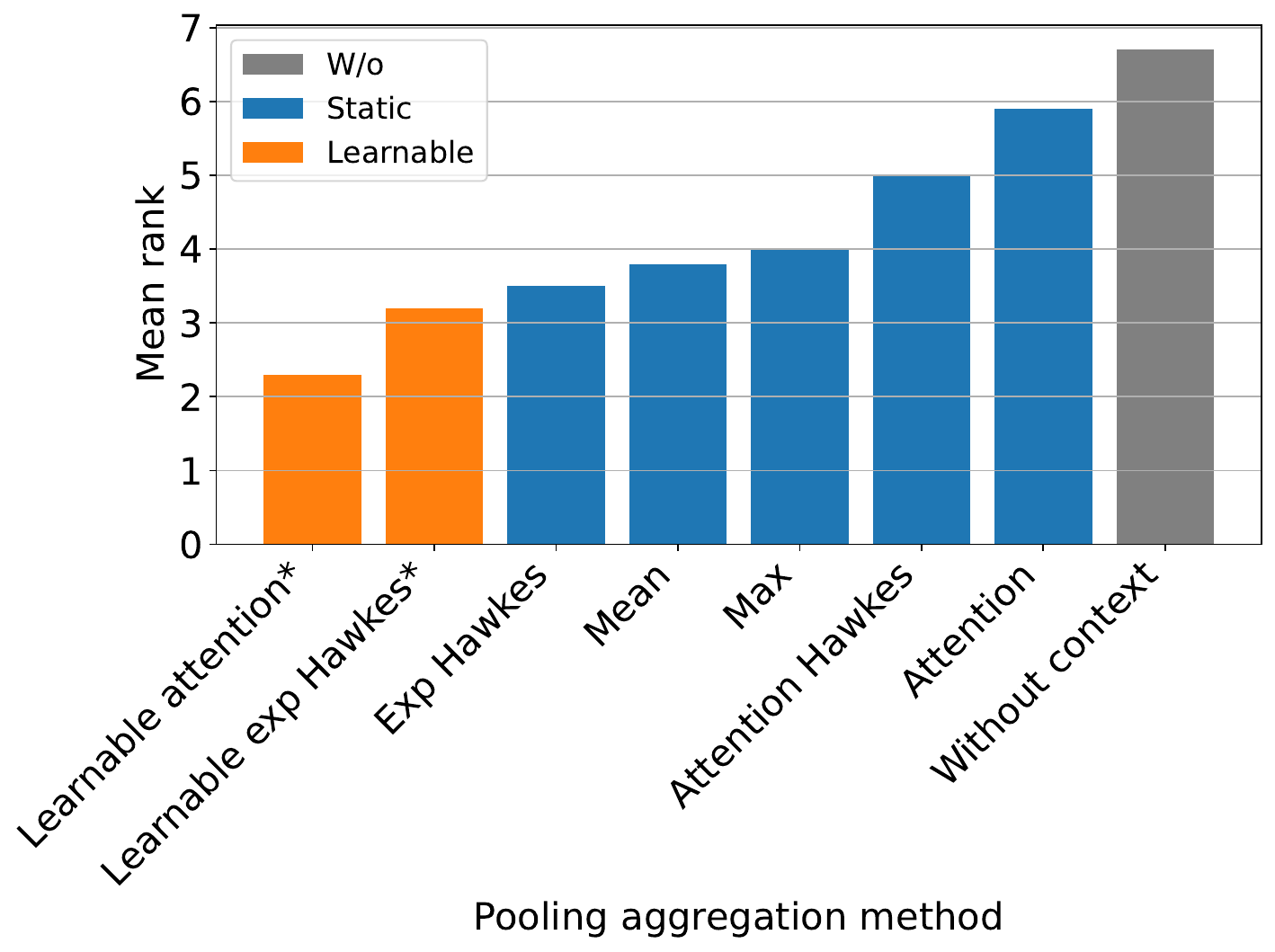}
     \caption{Mean rank ($\downarrow$) with respect to ROC-AUC values for methods of aggregating external context over nine datasets and two types of classification-based validation. The proposed Learnable attention approach has the lowest mean rank.}
     \label{fig:mean_rank}
\end{figure}

Given available large-scale datasets containing series with thousands of events, neural networks appear a reasonable methodological choice~\cite{shchur2021neural}.
Their strong performance has been confirmed for sequential data~\cite{babaev2019rnn}.
The main part of a neural network is an encoder.
It provides a representation vector for a sequence of events as input at a selected time moment, suitable for diverse downstream tasks~\cite{bin2022review}. 
Since most event sequences are unlabeled, self-supervised learning provides the most powerful representations~\cite{babaev2022coles}. 
With enhanced quality and robustness~\cite{bazarova2024universal}, these models are suitable for diverse applied tasks, including next event ~\cite{xue2023easytpp}, loan defaults prediction~\cite{zaytsev2023designing}, customer churn prediction, item recommendations, and fraud detection.

Processing sequences in isolation using a trained neural network constitutes an efficient and natural practice.
However, a small number of events per user and lengthy inactivity periods lead to a poor description of the current state. 
Thus, the representational power appears weak, despite the use of a strong model, due to data scarcity. 
With explicit manifestation of interdependencies across sequences, we can enrich the representation by aggregating signals across linked nodes, as temporal graph neural networks do.
The model performance improves~\cite{farajtabar2014shaping, farajtabar2017coevolve} if accounting for both the general context from recent events in connected sequences and the local one from similar sequences.
However, these approaches require affinity labels for pairs of sequences, which are typically unavailable for event sequence data.
A variant of temporal point-process-based models directly learns full interaction matrices \cite{hawkes2016de,passino2023mutually,shchur2021neural}.
This approach lacks scalability and a natural way to expand the user set, limiting its potential applications and causing stability issues. 
An alternative approach~\cite{bazarova2024universal} provides evidence that we can define a global external state by averaging across all users' representations. 
However, their method yields unstable quality improvements and fails to model complex interactions between sequences.

We propose a learnable sequence-embedding aggregation that provides an external context vector for a sequence, given a database of others' current states.
This approach enhances performance by incorporating additional information into the embeddings. 
The method works in the absence of direct affinity labels between sequences and naturally allows addition and removal of sequences from a database for aggregation and prediction.
In our \emph{Learnable attention}, the key novelty lies in leveraging a straightforward yet powerful and efficient attention-based pooling technique for external context aggregation by representation learning in sets of co-occurring event sequences.

Our main contributions are:
\begin{itemize}
    \item 
    \emph{A pioneering pipeline for enriching event sequence representation by aggregating representations of other sequences.} Our study explores methods to construct these aggregations using the same data and keeping the training and inference computationally efficient. The pipeline supports two modes: tuning a separate aggregation head or fine-tuning a whole aggregation-aware model.
    \item \emph{A novel Learnable attention aggregation method}, which stands out among the considered methods. It leverages a learnable attention mechanism to compute sequence similarities, providing weights for aggregation that focus on relevant sequences. 
    \item \emph{An empirical evaluation of external information aggregation} that spans over nine diverse datasets from different subject domains and problem statements, including global downstream tasks, next event type, and next item prediction, with two separate machine learning pipelines: a traditional sequential data processing pipeline (classification) and a recommendations pipeline (RecSys). Incorporating an external context vector improves performance across all scenarios, including fine-tuning the encoder or using a frozen backbone. Among the considered aggregation methods, the proposed \textit{Learnable attention} stands out, as evident from Figure~\ref{fig:mean_rank}.
    \item \emph{Empirical guidelines for efficient and robust external aggregation.} We find that the positive effect of aggregation is most evident post-distribution shift and that high performance can be achieved with under $1 000$ sequences, maintaining inference time overhead below $5\%$.
\end{itemize}

The code used to produce the results is available on our GitHub page: 

\href{https://github.com/petrsokerin/External-Context-Aggregation}{\url{https://github.com/petrsokerin/External-Context-Aggregation}}.

%% file: chapters_sigmod/02_literature.tex
\section{Related works}
\label{lit_rewiev}

We review the approaches to external context aggregation in various data modalities and methods for evaluating sequence representations. 
The review also considers existing representation learning methods for event sequences, including self-supervised methods for general people's behavior and for the bank transaction domain.

\subsection{Event sequence representation learning}
Neural networks have been shown to perform well when faced with the task of event sequence representation learning~\cite{babaev2022coles}.
Obtained representations are helpful both for tasks describing local in-time properties, such as next event prediction~\cite{zhuzhel2023continuous} or change point detection~\cite{ala2022deep,deldari2021time}, and whole sequence classification~\cite{bin2022review,jaiswal2020survey}.

\paragraph{Self-supervised learning (SSL)} SSL is a solid representation learning paradigm that learns an encoder --- a neural network to produce a representation vector without labeled data. 
This allows researchers to skip costly, possibly limited human expert annotation and to leverage large bodies of low-cost unlabeled data, leading to more universal representations~\cite{caron2021emerging}.
This paradigm is often implemented within contrastive and generative learning frameworks~\cite{liu2023ssl,zhang2020self}. 

In contrastive learning, the encoder learns to produce embeddings of objects (pictures, texts, audio, transactions, etc.). If two objects are similar, their embeddings should be close and vice versa.  
It originated in computer vision in the form of Siamese loss~\cite{hoffer2015deep} and SimCLR~\cite{chen2020simple} with subsequent DINO~\cite {caron2021emerging} and Barlow Twins~\cite{zbontar2021barlow} among others.
Contrastive learning also enables the generation of meaningful representations for time-series and event-sequence data~\cite{zhang2024self}.
Our work also uses a contrastive representation model, CoLES~\cite{babaev2022coles}, as it performs well on event sequence data and produces universal representations. 
The authors of the method studied various ways to define positive and negative pairs during training. 
They came up with the split strategy, which considers two subsequences from the same sequence as a positive pair and from different users as negative. 
There, representations of the subsequences were obtained via a Long Short-Term Memory network~\cite{hochreiter1997long} (LSTM) as an encoder, since transformers~\cite{vaswani2017attention} don't always perform better on sequential data, similar to~\cite{yue2022ts2vec}. However, our results indicate that there is no clear winner between LSTM and transformer encoders, as their performance varies depending on the specific task. According to prior work, the CoLES model is the state-of-the-art for processing transaction data~\cite {bazarova2025learning, fadeev2025latte}.  

Another powerful tool for constructing representation vectors can be Hawkes processes~\cite{hawkes2016de,hawkes1971spectra,hawkes2018hawkes,zhang2020self}. Hawkes processes, a class of self-exciting point processes, offer a powerful framework for modeling event sequences where the occurrence of one event increases the probability of subsequent events. This makes them well-suited for analyzing sequential data, where events such as purchases often exhibit temporal dependencies. 

\subsection{Accounting for external information}
Further improvement can be achieved by considering and aggregating the actions of other users~\cite{ala2022deep}.
It would allow us to model two effects: interactions between users and the information about the overall environment state (e.g., macroeconomic dynamics). 

For example, in the bank transaction domain, selecting appropriate macroeconomic indicators and accounting for them can improve model predictions. But the correct choice of these indicators requires specialized domain knowledge, often requiring input from a human expert. However, it has been shown that this information can be implicitly extracted using various methods to aggregate the actions of all bank clients (e.g., mean, max)~\cite{ala2022deep}. 
This approach allows the extraction of more information from the dataset in an end-to-end manner without additional annotation. 
Experiments in~\cite{bazarova2024universal} show that even naive aggregation within an external context provides superior results on some event-sequence modeling problems, though their results lacked stability.

One way to select data for aggregation is graph modeling. In graph representations, learning links between objects is a crucial component of a model, as implemented in state-of-the-art (SOTA) models~\cite{kipf2017semi}.
Similarly, this line of thought has developed in temporal point processes, particularly Hawkes mutually exciting point processes, where one can either derive or use provided node links~\cite{dizaji2022comparative,passino2023mutually}. 
However, in most cases, exact information between connections is absent, and their restoration is of limited quality~\cite{shumovskaia2021linking}.

Other examples of simultaneous accounting for different users' behavior and characteristics of a particular moment in time occur in recommendation systems. While classical algorithms consider only the similarity between users or items for recommendations, session-based~\cite{wang2021session} and time-based~\cite{ghiye2023timedecay,JAIN20231834,xia2010timedecay} recommendation systems integrate information from purchase flows across different users. Thus, we account for dynamic user preferences and ongoing external context.
In this case, the system balances user and group recommendations, as noted in a review~\cite{ceh2022performance}. 
Further papers included an attention mechanism in this workflow~\cite{guo2020group}. 
The limited scope of this work is constrained by efficiency considerations, which existing work addresses by providing recommendations for a predefined set of users.
However, a single sequence perspective can be crucial, given the diversity of possible life paths.

Recent advances in session-based recommendation further emphasize the role of cross-user context. For example, BERT4Rec~\cite{sun2019bert4rec} uses bidirectional self-attention over interaction sequences, allowing each event to incorporate both past and future context within a session, thereby yielding more informative representations. On the other hand, the SASRec~\cite{kang2018self} decoder-based transformer could remain a state-of-the-art model if it is trained with an alternative to the original paper loss function~\cite{klenitskiy2023turning, mezentsev2024scalable}. 

At the same time, Beyond Clicks~\cite{wang2020beyond} explicitly aggregates information across users by constructing a global item graph from multiple sessions, capturing shared transition patterns and heterogeneous user behavior. This indicates that cross-user signals can be incorporated not only through direct aggregation but also via structural representations. In large-scale industrial systems, this idea is often combined with self-supervised learning, as in~\cite{yao2021self}, where auxiliary objectives help uncover latent relationships between items and alleviate sparsity in user feedback. Finally, CSRM~\cite{jiang2025dynamic} incorporates cross-session information through an external memory module that aggregates representations from other users’ sessions, although its reliance on storing and retrieving similar sessions may limit scalability and reduce effectiveness in highly diverse or sparse settings. 

At the same time, existing approaches either use indirect ways to capture cross-user information or require more complex and less scalable designs. In contrast, our method leverages cross-user context in a more straightforward and efficient way. It remains lightweight and can be easily applied in large-scale settings, which highlights the practical value of our approach.

%% file: chapters_sigmod/03_methods.tex
\section{Methodology}\label{sec:methods}

\begin{figure*}[!t]
     \centering
     \includegraphics[width=0.9\textwidth]{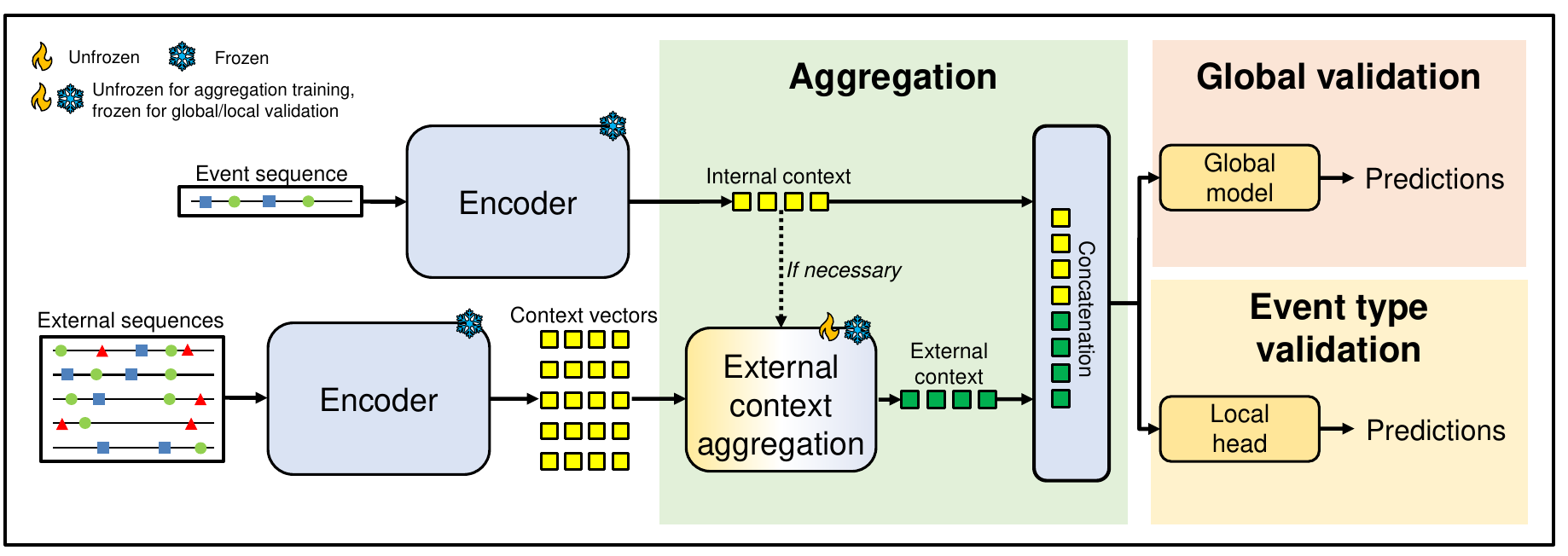}
     \caption{General classification pipeline for external context generation to integrate it with a vanilla internal context, followed by the considered validation procedures.}
     \label{fig:pipeline}
\end{figure*}

The general pipeline for using external information to solve a specific problem is depicted in Figure~\ref{fig:pipeline}. It has the following parts:
\begin{enumerate}
    \item Sequences of events are preprocessed and passed through a pretrained encoder to get embeddings.
    \item An external context aggregation method combines embeddings from different sequences and produces an external context vector at the current time moment. 
    \item The external context vector is concatenated with the embedding of the sequence under consideration. The concatenated vector replaces the original one, serving as the current representation of the sequence. It is used to solve a downstream task. 
\end{enumerate}

The corresponding subsections discuss in detail all the steps presented in the general pipeline.


\subsection{Event sequences data}

In this work, we consider datasets of event sequences.
Our sample $D = \{(s_i, y_i)\}_{i = 1}^N$, where $s_i$ is a sequence of events for user $i$ and $y_i$ is a target label for it.
Each event in a sequence has a vector description $\vecx$ and a timestamp $t$.
So, each sequence $s_i = \{(\vecx_{ij}, t_{ij})\}_{j = 1}^{T_i}$ with $T_i$ being the number of events and $t_{ij}$ are sorted in ascending order.


To standardize the pipeline for event sequences, we focus on two main features of each event, along with timestamps. In the financial domain, the transaction's merchant category code (MCC code) and the transaction's amount, which represents the amount of money spent in a transaction, are used as features. MCC code is the type of transaction, a categorical variable with $\sim 1000$ possible codes. For all other datasets, we also use categorical and numerical features, excluding Taobao, which contains only the MCC code as a feature. 
In sequential recommendations, the user's past interactions with items are also an event sequence $s_i = \{(\vecx_{ij}, t_{ij})\}_{j = 1}^{T_i} = \{(\mathbf{v}_{ij}, t_{ij})\}_{j = 1}^{T_i}$, where $v_{ij}$ is the item at position $j$ in the sequence, i.e., a categorical feature. 
So, if timestamps are available, the data structure is the same. 

Targets typically correspond to classification problems, including customer churn prediction, customer default, fraud detection, and gender prediction. In the RecSys scenario, we use a common downstream problem: ranking the next-item interactions. 




\subsection{An encoder architecture}
\label{sec:architecture}

Given a sequence $s_i$, the encoder produces the matrix $ (\mathbf{h}_{i, 1}, \ldots, \mathbf{h}_{i, T_i})$ from $\mathbb{R}^{T_i \times m}$ corresponding to a sequence with length $T_i$ with the embedding dimension $m$.
These are the \emph{internal representations}.
For a specific time moment $t$, the internal representation is the last available vector $\mathbf{h}_{i, t} = \mathbf{h}_{i, j(t)}$ from the past with $j(t) = \mathrm{arg} \max_k t_k$, such that $t_k \leq t$. All representations are computed using only events observed up to time t, ensuring temporal causality.

For different data types, we use established architectural and design choices. 

For all datasets excluding RecSys, our encoder includes the input layers that transform features for each timestamp, followed by the subsequent encoder body. 
For a categorical feature, we produce a learnable dictionary-based embedding~\cite{pennington2014glove} of dimension $d_{\mathrm{cat}}$; it is then concatenated with $d_{\mathrm{cont}}$ normalized continuous features.
Thus, as the input to the encoder body, we have a sequence of the dimension $T_i \times (d_{\mathrm{cat}} + d_{\mathrm{cont}})$.
We take an RNN-based or a Transformer-based architecture as a model backbone.
Depending on the dataset, the RNN block in the RNN-based backbone is either a Long Short-Term Memory (LSTM) or a Gated Recurrent Unit (GRU). For the transformer model, we use self-attention bidirectional transformer blocks. The baseline encoder architecture is pretrained within the self-supervised CoLES framework~\cite{babaev2022coles}.

For the recommendation system pipeline, we use a transformer-based model called SASRec~\cite{kang2018self}. This model’s architecture is similar to that of transformer decoders, such as GPT. Recommender systems traditionally solve a ranking task, so we need to compute the relevance score $r_{ij}$ between user $i$ and item $j$ to select the top-$K$ items for each user. To compute relevance scores for all candidate items, we calculate the dot product between the user embedding $\mathbf{u}_i$ and matrix $E$, where $E$ is the item embedding matrix from $\mathbb{R}^{L \times m}$ and $L$ is the number of unique items. In SASRec, the model receives sequences of user events as input and produces a sequence of hidden states as output; the user representation is taken to be the last hidden state, $\mathbf{u}_i = \mathbf{h}_{i, T_i}$. More formally, the relevance vector for user $i$ is computed as $\mathbf{r}_i = E \mathbf{u}_i = E_i \mathbf{h}_{i, T_i}$.
Unlike models such as GPT, which use a separate linear head for final predictions, SASRec utilizes the same item embedding matrix both as input to the transformer and for generating the final relevance scores.

\subsection{Representation-based aggregation of external information}
\label{sec:global_context_methods}

A trained encoder produces representations for all sequences in the sample $D$, defining the internal representation vector $\vech_{i, t}$ for each user $i$ and time moment $t$. This vector may not contain enough information about the user's behavior, so external factors can improve performance on downstream tasks. We propose obtaining these factors from an additional external context representation vector constructed by aggregating the internal representation vectors of a random subset of $n$ selected users. 

\begin{figure}[b]
     \centering
     \includegraphics[width=0.85\columnwidth]{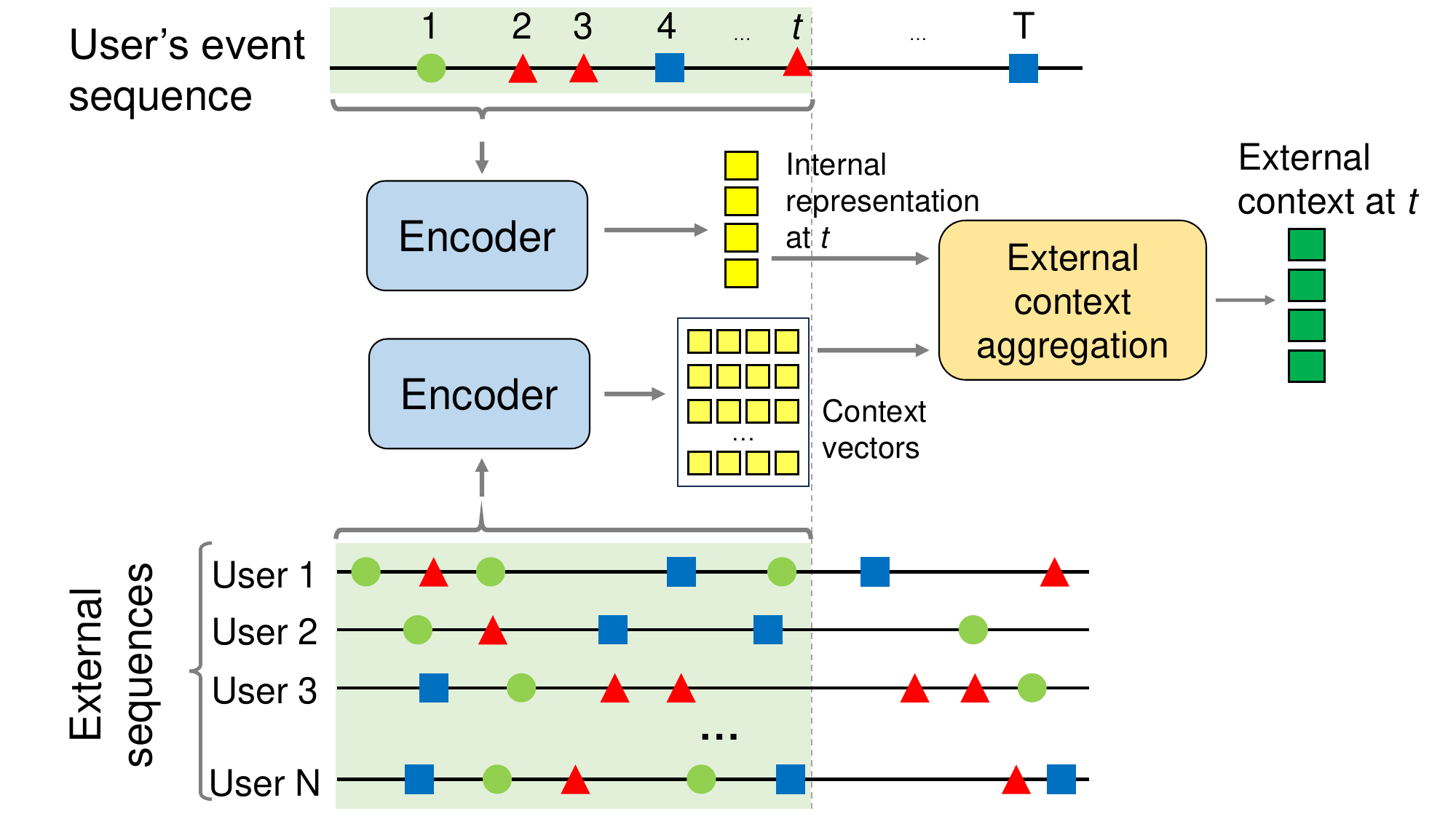}     
     \caption{External context aggregation that outputs the external representation. Here, we don't show the encoding of external sequences.}
     \label{fig:global_pooling}
\end{figure}

The procedure for constructing an external context vector at a specific point in time $t$ is presented in Figure~\ref{fig:global_pooling} and is as follows:
\begin{itemize}
     \item [1.] Obtain an internal representation $\mathbf{h}_t$ of a specific user at time~$t$. We use the last available representation from the past. Our goal is to construct an external context vector for this user.
     \item [2.] Select internal representations from $n$ other selected users observed shortly before the current time point $t$: 
     \begin{equation}
     H_t = \{\mathbf{h}_{i, t} \}_{i = 1}^n. 
     \end{equation}
     $H_t \in \mathbb{R}^{m \times n}$ is a matrix, whose columns are embeddings of size $m$ for $n$ users from dataset at a given time point $t$. 
     \item [3.] Apply an aggregation $\mathcal{A}(\cdot, \cdot)$ to the resulting set of vectors given the current embedding vector for a specific user $\mathbf{h}_t$: 
     \begin{equation}
      \mathbf{g}_t = \mathcal{A}(H_t, \mathbf{h}_t),
     \end{equation}
     where the resulting vector $\mathbf{g}_t$ is the vector of external context, that we call \emph{external representation}.
\end{itemize}

\begin{figure}[!t]
     \centering
     \includegraphics[width=0.99\columnwidth]{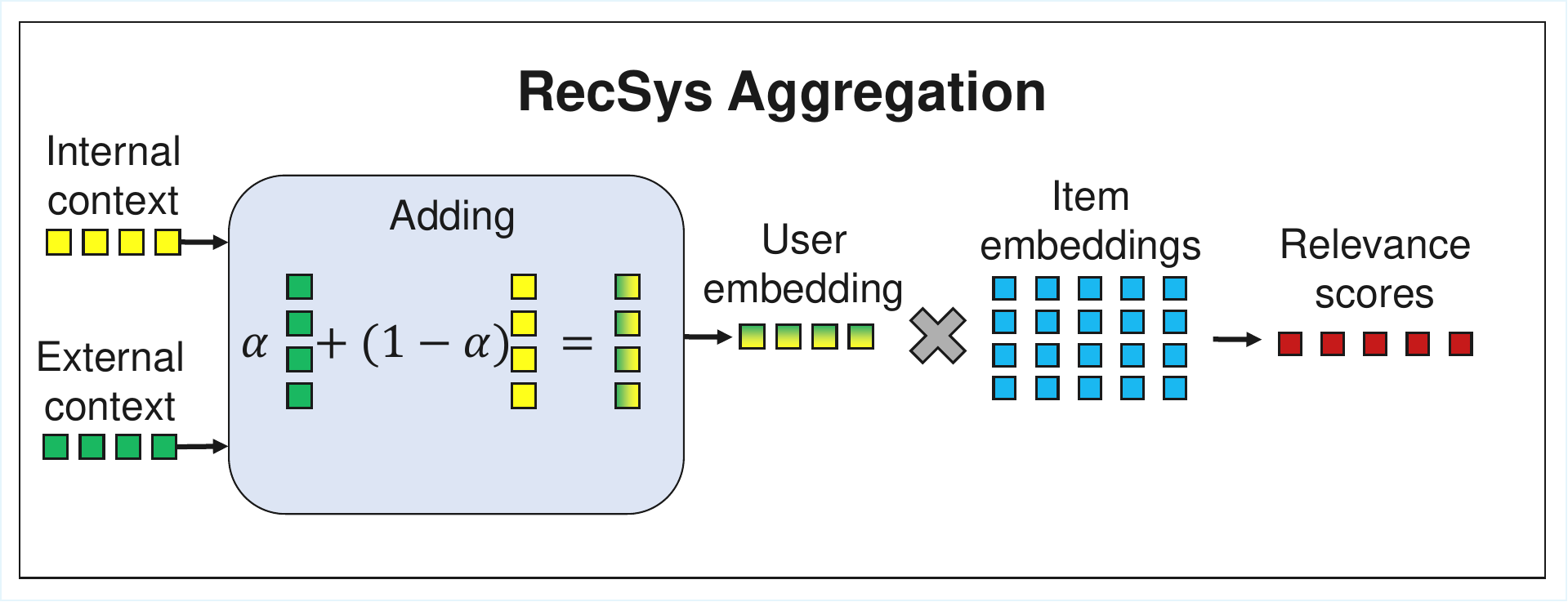}
     \caption{General RecSys pipeline for external context integration with internal context.}
     \label{fig:recsys_pipeline}
\end{figure}

To solve the applied problem, we concatenate the resulting context embedding vector with the user’s internal embedding and validate the extended representation $(\vech_t, \vecg_t)$ when it is possible. For the SASRec architecture, we can't concatenate the external and internal feature vectors. In this case, the final prediction should be computed as the scalar product of the item embedding matrix with a fixed embedding size $m$. We first train this model without an external context as a pre-training stage, then use it to obtain external vectors. After getting an updated user feature vector $\tilde{\mathbf{u}}_i$ as a weighted sum of vanilla SASRec (internal context) user vector and external context vector  $\tilde{\mathbf{u}}_i = (1 - \alpha) \mathbf{u}_{i} + \alpha \mathbf{g}_{i, T_i} = (1 - \alpha)\mathbf{h}_{i, T_i} + \alpha \mathbf{g}_{i, T_i}$, where $\alpha$ is a weight coefficient hyperparameter as presented in Figure~\ref{fig:recsys_pipeline}. 

\paragraph{$N, n$ correspondence}
We perform aggregation over $n$ randomly sampled users instead of all $N$.
It makes our method more scalable for datasets with a large number of sequences $N$.
Our experiments support the evidence that it is unnecessary to use all users for aggregation, and increasing $n$ above a certain value results in diminishing returns for quality. 

\subsection{Considered aggregation methods}

The aggregation function $\mathcal{A}(H_t, \mathbf{h}_{i, j(i)})$ in most cases has the form: 
\begin{equation}
\mathbf{g}_t = \mathcal{A}(H_t, \mathbf{h}_t) = \sum_{i=1}^n w_i \mathbf{h}_{i, t},
\end{equation}
where the weights $w_i$ of the $i$-th user in the aggregation depends on correspondence between $\mathbf{h}_t$ and $H_t$, and interaction within these vectors. In some cases, the aggregation function can have trainable parameters.
These functions are trained in a self-supervised manner for vectors $\mathbf{g}_t$ as embeddings with the contrastive~\cite{babaev2022coles} or generative next-event-type prediction (for non-RecSys datasets) or full cross-entropy (for RecSys datasets) loss function; this procedure shares the loss function with the original encoder.
Thus, our approach remains self-supervised.

\subsubsection{Straightforward aggregation}

Mean and Max pooling provide a natural way to aggregate external information.
For Mean, $\mathbf{g}_t = \frac{1}{n} \sum_{i = 1}^n \mathbf{h}_{i, t}$.
For Max, the maximum value for each component over $n$ vectors is taken.
Here, we adopt the concepts of Mean and Max Pooling operations from convolutional neural networks~\cite{boureau2010theoretical} and language models~\cite{xing2024comparative}, now applied to a set of embeddings from selected users.

\subsubsection{Methods based on the attention mechanism}

The pooling aggregations introduced above produce similar $\mathbf{g}_t$ for all users. 
Thus, they ignore interactions between users from the dataset.

Given similarities among users, one can adopt aggregation-based approaches from spatial graphs~\cite{huang2024temporal}, weighting the vectors $\mathbf{h}_{i, t}$ according to the similarities between the sample users and a selected user.
User similarities are absent in the data in most cases, and we can't use common GNN techniques, so we infer them by comparing an internal representation $\mathbf{h}_t$ and selected representations $H_t$.  

The attention mechanism~\cite{vaswani2017attention} is natural for such a problem. 
We compare different adoptions of attention.

\textbf{Simple attention without learnable parameters.}
The simplest attention-based solution is the following:
\begin{equation}
     \vecg_t = H_t \mathrm{softmax} (H_t^T \vech_t).
\end{equation}
In this case, the user similarity metric is normalized by the softmax of the dot product to produce similarity-based weights for vectors from $H_t$.

\textbf{Learnable attention.}
We propose a Learnable attention method that combines a simple attention mechanism with learnable parameters. Basically, we can define the method as follows:

\begin{equation}
     \vecg_t = H_t \mathrm{softmax} (w_{\theta}(H_t^T, \vech_t)),
\end{equation}
where $w_{\theta}$ is an attention-based function with learnable parameters. 

The simplest way to define this function is to use a trainable matrix $A$:
\begin{equation}
     \vecg_t = H_t \mathrm{softmax} (H_t^T A \vech_t),
\end{equation}
where $A \in \mathbb{R}^{m \times m}$ is the matrix to be trained.
In this method, called \textit{Matrix attention}, before calculating the scalar product, vectors of representations from the dataset are passed through a trainable linear layer. 

Another method, called \textit{Kernel attention}, applies the idea that we can use the general kernel to calculate attention scores:

\begin{equation}
     \vecg_t = H_t \mathrm{softmax} (\langle \phi(H_t), \phi(\vech_t) \rangle).
\end{equation}
There, $\phi(\cdot)$ is a learnable function applied to a row or independently to all rows if the input is a matrix, passed to a scalar product operator. In our case, it is parameterized by a two-layer fully-connected neural network with hidden size $c$.

\subsubsection{Methods inspired by Hawkes process}
The above methods ignore the last event time when defining similarities,
while it is natural to reduce similarities if the considered events are distant in time.
To validate or invalidate the importance of this information for the quality of the embeddings produced, we present Hawkes-process-inspired aggregation methods.
The exact mechanism and corresponding weights arise when we assume specific conditional intensities of events that depend on the past.
Specifically, Multivariate Hawkes processes ~\cite{hawkes1971spectra,hawkes2018hawkes} incorporate such dependencies, providing a generalization of the Poisson point process with the intensity that is conditional on past events.


The original Hawkes process is described by the following formula for the conditional intensity function $\lambda_u(t)$:
\begin{equation}
\lambda_u(t) =
\mu_u(t) + \sum_{v} \sum_{e_i \in \mathcal{H}_v(t)} b_{uv}  \kappa(t-t_i),
\end{equation}
where the first term, $\mu_u(t)$, models the current event sequence behaviour, and the second term, with $b_{uv}$, models interaction between sequence $u$ and sequence $v$,  $\kappa(t - t_i)$ is a kernel function of time, usually exponential, between time $t$ and time $t_i$ of event $e_i$. $e_i \in \mathcal{H}_v(t)$ are all events from sequence $v$ before time $t$. This expression defines the current event rate for sequence 
$u$ as the sum of a background rate and the influence of all past events from all sequences, where each event adds a temporary increase to the rate via a kernel function. In other words, the Hawkes process models how past events excite future events, with earlier events receiving less weight.

In the original Hawkes process, events from a single sequence have the same type. In our case, this is not true. Also, as we work not with original sequences but with their representations, we can simplify the equation:
\begin{equation}
\lambda_u(t) = \vecmu_u + \sum_{v} b(\vecmu_u, \vecmu_v) \kappa(t - t_v).
\end{equation}
Here, $\vecmu_u$ is a representation vector of the sequence under consideration, $\vecmu_v$ is a representation vector of the sequence from the training dataset,  and $t_v$ is the time moment of the last event in the $v$ sequence at the current time point, and $b$ and $\kappa$ are learnable kernels. The second term in the equation is the target external representation. 

In our notation, the equation can be rewritten:
\begin{equation}
     \mathbf{g}_t = b(H_t, \mathbf{h}_t)\cdot \kappa(t \cdot \mathbf{1} - \mathbf{T}),
\end{equation}
where $\mathbf{T}$ is a vector of  last event times in $H_t$ for the current time~$t$.

We simplify this general formulation to a set of different special cases. 
Mostly, this step is taken due to the instability of learning in a general case.
The different variations of the presented formulation of the problem are enumerated below.

\textbf{Exponential Hawkes}. In this simple variation of the Hawkes method, we use an exponential kernel for time transformation and an identical transformation for the matrix $X$.
\begin{equation}
     \mathbf{g}_t = H_t \exp{(-\gamma(t \cdot \mathbf{1} - \mathbf{T}))}.
\end{equation}

In this case, we weigh the vectors from $H$ with exponential time-dependent weights. In this method and the methods below, we use a weight coefficient $\gamma$ as a hyperparameter rather than the kernel function. This is necessary because time is conventionally encoded as a timestamp, which can exceed values of one million. Large time differences can cause numerical instability when passed through the exponential ($\exp$) operator.

\textbf{Learnable exponential Hawkes}

The learnable transformation was used in this method. Firstly, we concatenate each vector from  $H$ with $\mathbf{h}_t$ and get matrix $\tilde{H_t} \in \mathbb{R}^{2m \times n}$. Secondly, we pass the concatenated matrix through the feed-forward neural network $\phi_{NN}(\cdot)$ and get the matrix  $H_t'\in \mathbb{R}^{m \times n}$. 


\begin{equation}
\begin{split}
    \mathbf{g}_t &= \phi_{\mathrm{NN}}(\mathrm{concatenate}(H_t, \mathbf{h}_t))\exp{(-\gamma(t \cdot \mathbf{1} - \mathbf{T}))} \\
    &= \phi_{\mathrm{NN}}(\tilde{H_t})\exp{(-\gamma(t \cdot \mathbf{1} - \mathbf{T}))} \\
    &= H_t'\exp{(-\gamma(t \cdot \mathbf{1} - \mathbf{T}))}.
\end{split}
\end{equation}
Here, we additionally consider the dependencies between the current embedding vector and embedding vectors from the dataset.

\textbf{Attention Hawkes}. In this method, we combine the usual exponential Hawkes and the attention methods:
\begin{equation}
     \mathbf{g}_t = H_t\mathrm{softmax}(H_t^T \mathbf{h}_t)\exp{(-\gamma(t \cdot \mathbf{1} - \mathbf{T}))}.
\end{equation}

This approach has double weighting: the first accounts for user similarity, and the second accounts for the time delta.

The proposed methods above have two main advantages: they are lightweight and applicable to event sequences without additional cross-user interaction data. More complex architectures, such as graph neural networks or full cross-attention transformers, could theoretically be considered for external aggregation, but they face practical limitations in our setting. Graph-based approaches require explicit or reliably inferred relational structures between sequences, which are typically absent in event sequence data. Likewise, heavy transformer variants introduce substantial computational and memory overhead during inference, contradicting our emphasis on lightweight, production-ready aggregation. 

\subsection{Efficient work in production}
\label{sec:efficient_work}

To integrate our external-aggregation method into the production system for inference, we maintain a database of embedding vectors that enables on-the-fly aggregation.





The vector database maintains both internal and external representation vectors for each selected user for aggregation. As a database, we can take a variety of databases that can store float features.
The internal representation vectors are updated whenever a user performs a new action. In the case of simple aggregations shared across all users—such as \emph{Max} and \emph{Mean} --- the external aggregation vectors can be updated simultaneously. 
For more complex aggregation that involves individual aggregation vectors per user, we propose avoiding a full database update for each event. Instead, the database is updated at fixed time intervals. The duration of these intervals depends on the specific dataset, particularly the average time between user actions.
For instance, in the Churn dataset, a typical user performs no more than three transactions per day. Thus, a daily update frequency is a reasonable choice for maintaining the database.

This method ensures linear computational complexity in the number of users in the dataset. When aggregation is performed over a bounded subset of users ($n \ll N$), the additional inference cost scales linearly in $n$ and remains small relative to encoder computation. Consequently, our approach is both straightforward to implement and efficient to deploy in real-world environments, introducing minimal computational overhead.

\subsubsection{Computational complexity}

With precomputed external representations $H_t\in\mathbb{R}^{n\times m}$ and some additional matrices for the corresponding methods, aggregation adds the following incremental costs (Table \ref{tab:inference_cost}). Since the encoder cost $O(Tm^2)$ dominates, where $T$ is the length of the sequence, aggregation adds at most $O(mn+m^2)$ (for kernel variants) with modest constants, the extra inference cost is negligible for practical sizes (e.g., $m\approx 10^3$ and $n\leq 10^3$). The training-time overhead of learnable aggregations scales to $O(m^2 n)$ in the worst case and remains comparable to the baseline training time. 
This behavior is also empirically confirmed in our timing measurements from Table~\ref{tab:time}, presented in ~\ref{sec: res_time} section. 

\input{tables/inference_cost}

\subsection{Theoretical justification for external aggregation}

We consider a stationary zero-mean Gaussian process-based probabilistic data model~\cite{williams2006gaussian}. 
It considers a general family of functions that follow this assumption, as this family is dense in the space of continuous functions.
Moreover, it has a natural way to interpret external context through co-kriging~\cite{kennedy2001bayesian, zaytsev2016reliable}, where we have a set of dependent signals, and we can use their correlations to improve overall predictions.
While in some cases these models are mentioned in analytical solutions~\cite{zaytsev2017minimax,wendland2004scattered}, in our case, connected to extrapolation, it is unknown how to obtain forms suitable for analytical analysis. 
Instead, we conduct numerical experiments that provide a general picture of the usefulness of external context.

\subsubsection{Probabilistic model}

Here and below, notation is adopted from~\cite{zaytsev2017minimax} to describe a natural model with internal and external signals that, combined, can reduce the prediction error.
We consider a series of realizations of one-dimensional Gaussian (GP):
\begin{equation}
	x_i(t) = z_i(t) + y(t) + \varepsilon_i(t), i = 0, \ldots, K.
\end{equation}
Here, each $z_i(t) \sim \mathrm{GP}(0, k)$, zero-mean Gaussian process with the covariance function $k$,
$y(t) \sim \mathrm{GP}(0, k)$ is another zero-mean Gaussian process representing a \emph{trend} with the covariance function $k$,
and $\varepsilon_i(t)$ is a Gaussian white-noise process with variance $\sigma^2$. 
These separate processes are independent of each other.

Without loss of generality, we construct BLUP (best linear unbiased predictor) $\hat{x}_0(t) = \hat{x}(t)$ for $x_0(t) = x(t)$ that uses all available observations from all GPs. 

For the Gaussian model under consideration, the BLUP predictor coincides with the
conditional expectation:
\begin{equation}
\hat{x}_0(t)=\mathbb{E}[x_0(t)\mid X_D],
\end{equation}
where $X_D$ is a training sample.

For a specific training sample $D$, we define the quadratic~risk:
\begin{equation}
R(D, h, \delta) = \mathbb{E} \int_{h}^{h + \delta} (\hat{x}(t) - x(t))^2 dt,
\end{equation}
Using the standard Gaussian process posterior variance formula (see Appendix~\ref{sec:theory}), we obtain
\begin{equation}
R(D,h,\delta)
=
\int_h^{h+\delta}\operatorname{Var}(x_0(t)\mid X_D)\,dt,
\end{equation}
where $h$ controls the interval shift and $\delta$ controls the interval length; the expectation is taken with respect to realizations of GPs at points from $D$.
Our goal is to make the quadratic risk as small as possible using available data.

We work in the extrapolation regime and, without loss of generality, assume that all observations are to the left of $0$.
Also, to simplify analysis, we consider observations on a grid with a few last $m$ points missed:
\begin{equation}
	D_i^{m, \delta} = \{(-j\delta, x_i(-j\delta))\}_{j = -m}^{-\infty}.
\end{equation}
For $i = 1, \ldots, K$ we would use $m = 0$, for $i = 0$ we would vary $m$ to imitate last $m$ missing observations.
So, $D_K = D_K^{m, \delta} = D_0^{m, \delta} \cup \left( \cup_{i = 1, \ldots K} D_i^{0, \delta} \right)$. 
Also, we can't use an infinite sample size in numerical experiments.
Instead, we take a large enough value, since, for most covariance functions, the dependence on past observations decreases exponentially with increasing distance from the prediction~point.

In the introduced probabilistic model, $x_0(t)$ is the target to predict, and its observations constitute an internal context; other $x_i(t)$ provide an external context that helps to uncover the trend and thus reduce the quadratic error, introduced as the risk of integrated extrapolation $R(D, h, \delta)$.

\subsubsection{Covariance functions}

We use standard stationary covariance functions with distinct behavior patterns: exponential and squared exponential~\cite{williams2006gaussian}.
The exponential covariance function has the following form:
\begin{equation}
	k(t, t') = k(t - t') = \exp \left(-|t - t'| \right).
\end{equation}
The squared exponential covariance function has the following form:
\begin{equation}
	k(t, t') = k(t - t') = \exp \left(-(t - t')^2 \right).
\end{equation}

For the white noise process, we would also fix some noise variance $\sigma^2$ in the experiments to completely define it.

\subsubsection{Numerical experiments}

For each covariance function, we vary three components --- $m$, $K$, and the relative weight of the shared trend --- to examine the effect of external information on the results:
\begin{itemize}
    \item In the first experiment, our goal is to examine how the usefulness of external information depends on the contribution of the shared trend.
    For this purpose, we consider the weighted model
    \begin{equation}
        \label{eq: teor_weight}
        x_i(t) = (1-w) z_i(t) + w\, y(t) + \varepsilon_i(t),
    \end{equation}
    where $w \in [0,1]$ is the trend weight.
    We then evaluate the integrated risk as a function of $w$ for several values of $K$, showing how the gain from external observations changes as the common trend becomes more~pronounced.
    \item In the second experiment, our goal is to see how missed observations for the core $x_0(t)$ affect the added effects of data from other series.
    We fix a small $K = 5$ and start to increase $m$, obtaining the curve for the integrated risk $R(D, h, \delta)$ on a grid of integer $\delta$ values.
    
    \item In the third experiment, our goal is to see what value of $K$ is enough.
    We consider various options for $m$ and see how the error changes if we increase $K$ from $0$ to $50$, considering a non-uniform grid with denser integer values of $K$ in the beginning, obtaining curves of $R(D_K, h, \delta)$ for different $K$ values.

\end{itemize}

Technically, we use the same covariance functions for trend and distinct covariance functions. 
In practice, the quality improvement due to the external information would naturally depend on them and on relative weights for $z_i$ and $y$ in $x_i$.

\paragraph{Effect of the trend weight.}
To additionally study how the usefulness of external context depends on the strength of the shared component, we consider the following weighted probabilistic model, presented in Equation~\ref{eq: teor_weight}.
For small values of $w$, the individual component $z_i(t)$ dominates, so different series are only weakly dependent.
For large values of $w$, the common trend $y(t)$ contributes more strongly, which increases cross-series dependence and makes external observations more informative for predicting the target process.

\begin{figure}[t]
    \centering
    \begin{subfigure}[t]{0.42\textwidth}
        \centering
        \includegraphics[width=\textwidth]{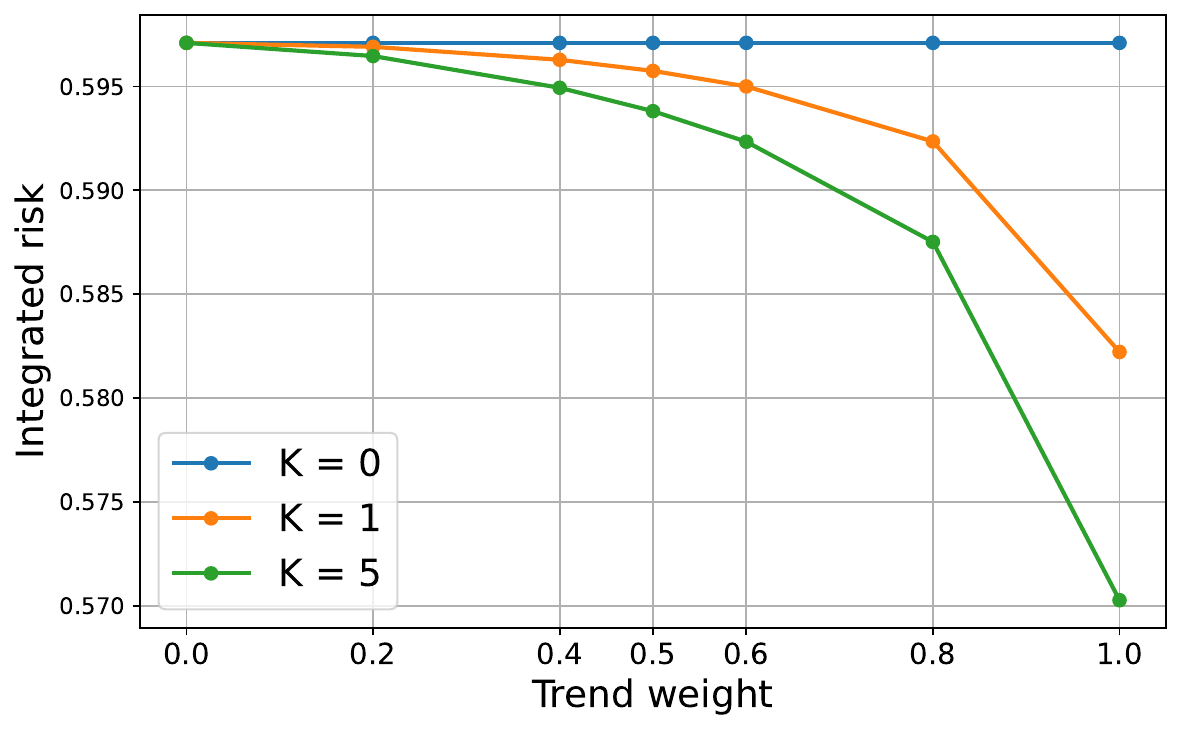}
        \caption{Exponential covariance function.}
        \label{fig:risk-vs-trend-weight-exp}
    \end{subfigure}
    \hfill
    \begin{subfigure}[t]{0.42\textwidth}
        \centering
        \includegraphics[width=\textwidth]{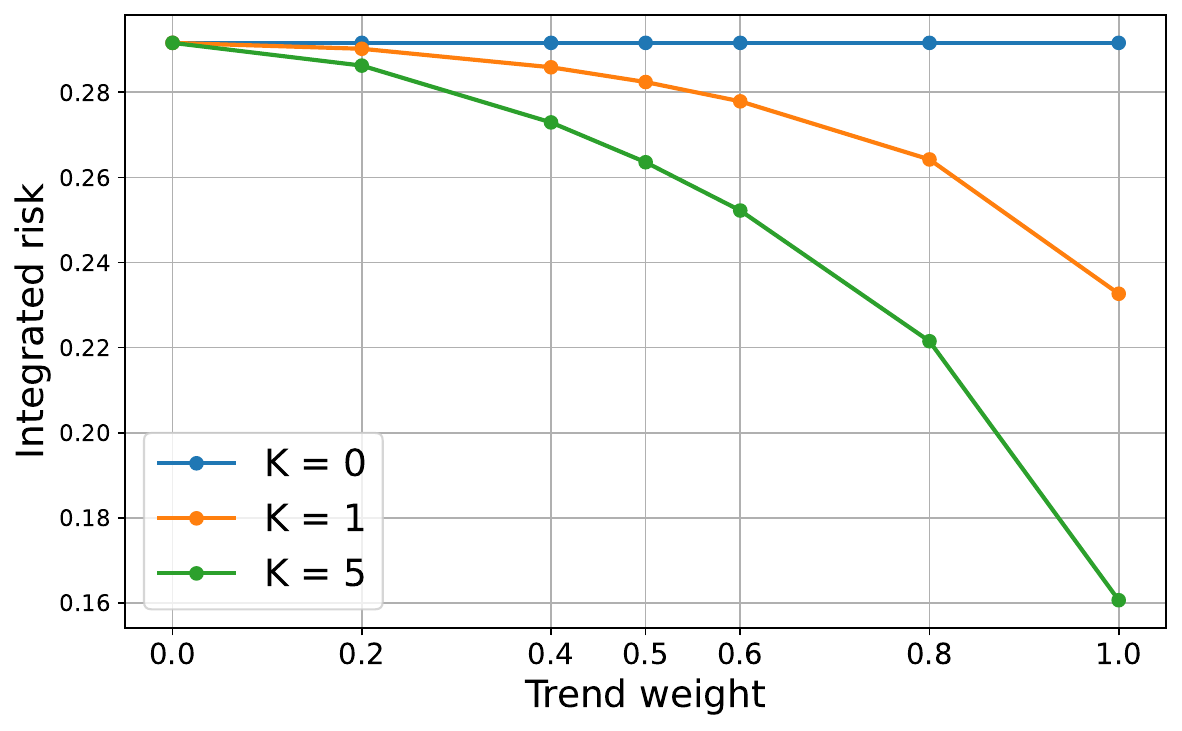}
        \caption{Squared exponential covariance function.}
        \label{fig:risk-vs-trend-weight-se}
    \end{subfigure}
    \caption{
    Integrated extrapolation risk (lower is better) as a function of the trend weight $w$ in the model
    $x_i(t) = (1-w) z_i(t) + w\, y(t) + \varepsilon_i(t)$.
    Different curves correspond to different numbers of external series $K$.
    }
    \label{fig:risk-vs-trend-weight}
\end{figure}

In this experiment, we fix the remaining parameters and evaluate the integrated extrapolation risk as a function of $w$ for several values of $K$.
The results, shown in Figure~\ref{fig:risk-vs-trend-weight}, confirm the expected behavior:
as the trend weight increases, the gain from external information becomes more pronounced.

\begin{figure}[t]
    \centering
    \begin{subfigure}[t]{0.42\textwidth}
        \centering
        \includegraphics[width=\textwidth]{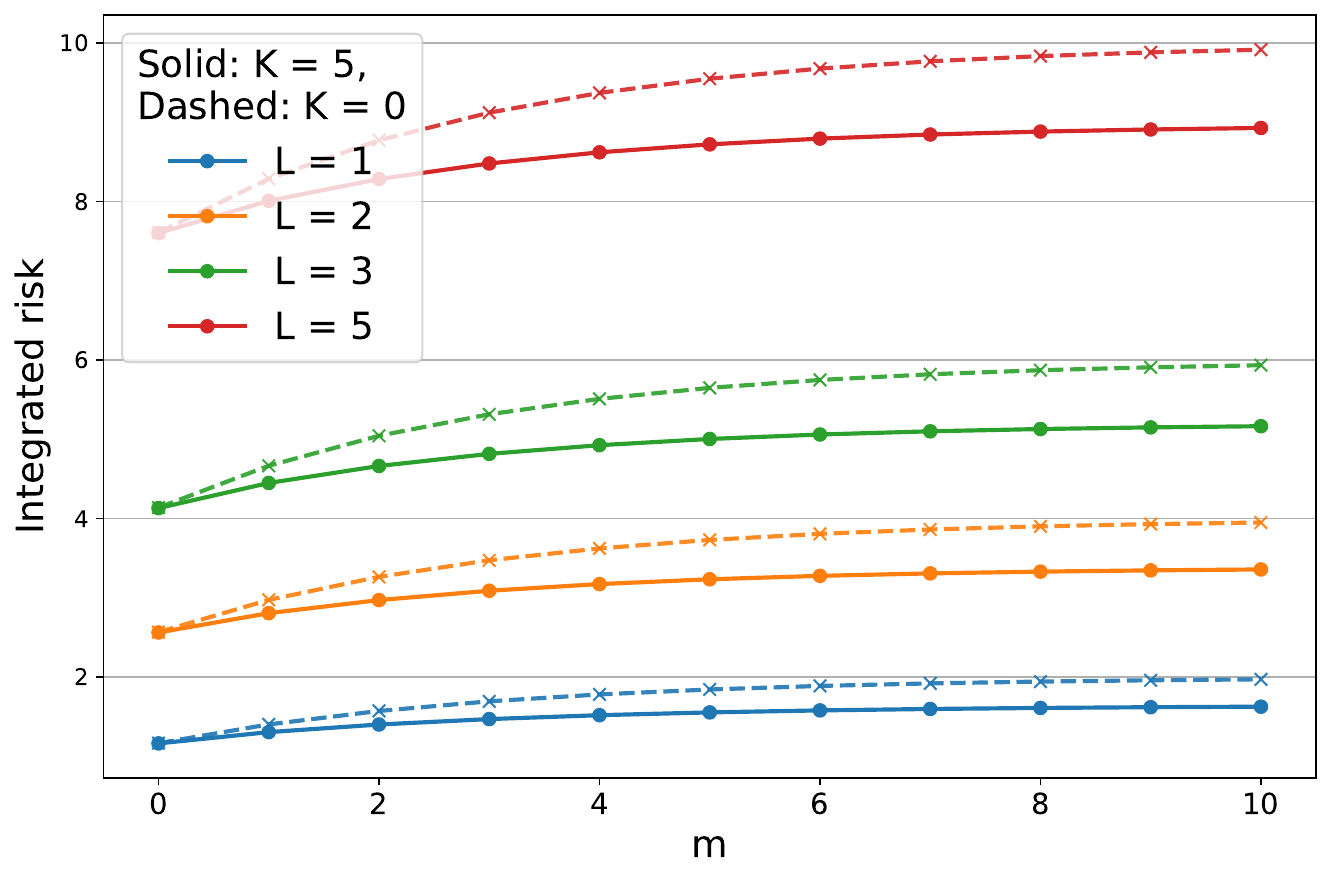}
        \caption{Exponential covariance function.}
        \label{fig:risk-vs-m-exp}
    \end{subfigure}
    \hfill
    \begin{subfigure}[t]{0.42\textwidth}
        \centering
        \includegraphics[width=\textwidth]{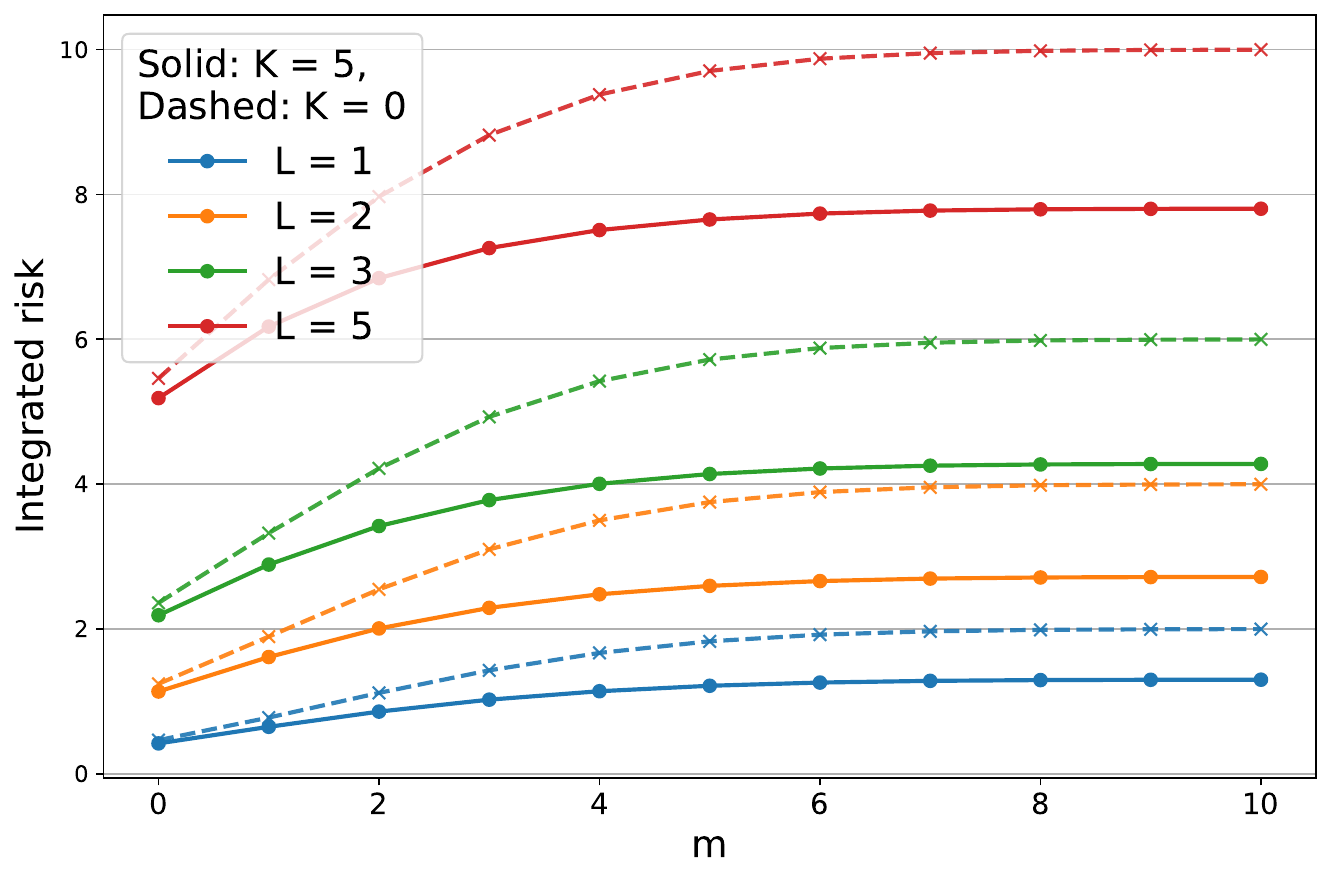}
        \caption{Squared exponential covariance function.}
        \label{fig:risk-vs-m-se}
    \end{subfigure}
    \caption{
    Integrated extrapolation risk $R(D_K, 0, L)$ (lower is better) as a function of the number of missing recent observations $m$ of the target process for the exponential and squared exponential covariance functions. 
    Different curves correspond to different prediction interval lengths $L$ and to the use ($K=5$) or absence ($K=0$) of external observations. 
    }
    \label{fig:risk-vs-m}
\end{figure}

\paragraph{Effect of missing observations.}
The plot for natural parameters of Gaussian processes is presented in Figure~\ref{fig:risk-vs-m}.
We see that the risk reduction is substantial across all considered intervals in the considered settings. 
The gains for squared exponential covariance function models are more evident, as they are more affected by the absence of recent observations.

\begin{figure}[t]
    \centering
    \begin{subfigure}[t]{0.42\textwidth}
        \centering
        \includegraphics[width=\textwidth]{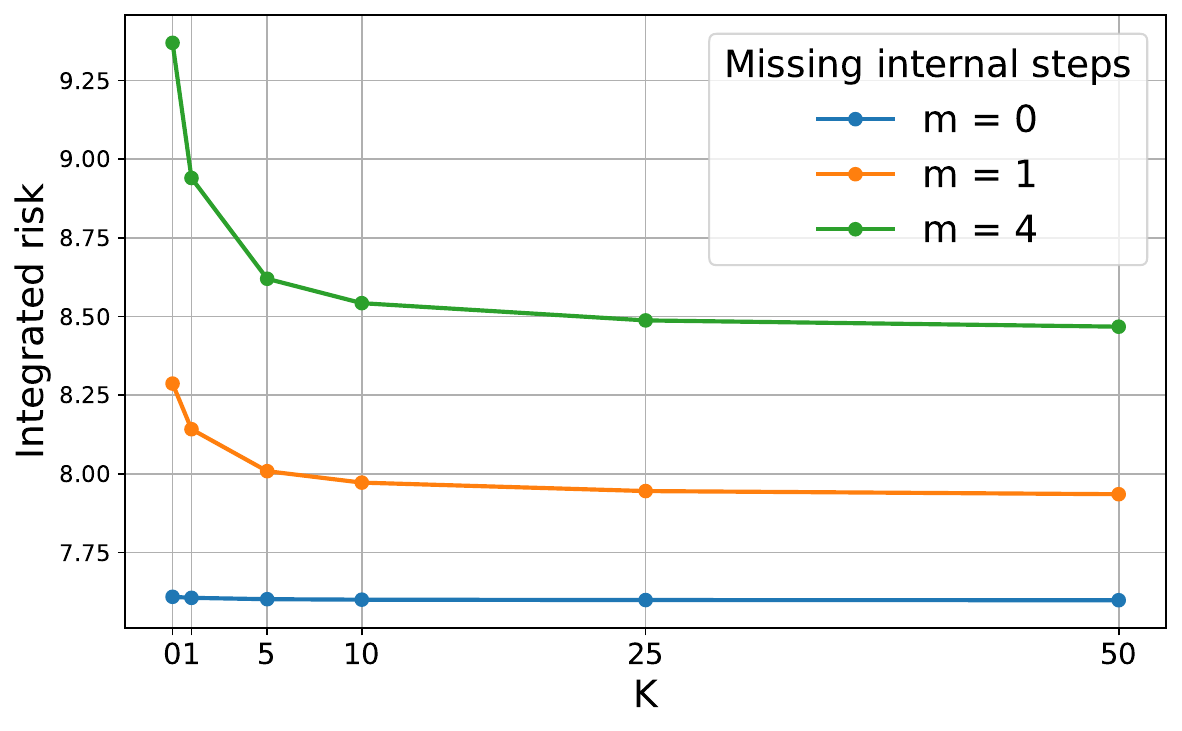}
        \caption{Exponential covariance function.}
        \label{fig:risk-vs-K-exp-L5}
    \end{subfigure}
    \hfill
    \begin{subfigure}[t]{0.42\textwidth}
        \centering
        \includegraphics[width=\textwidth]{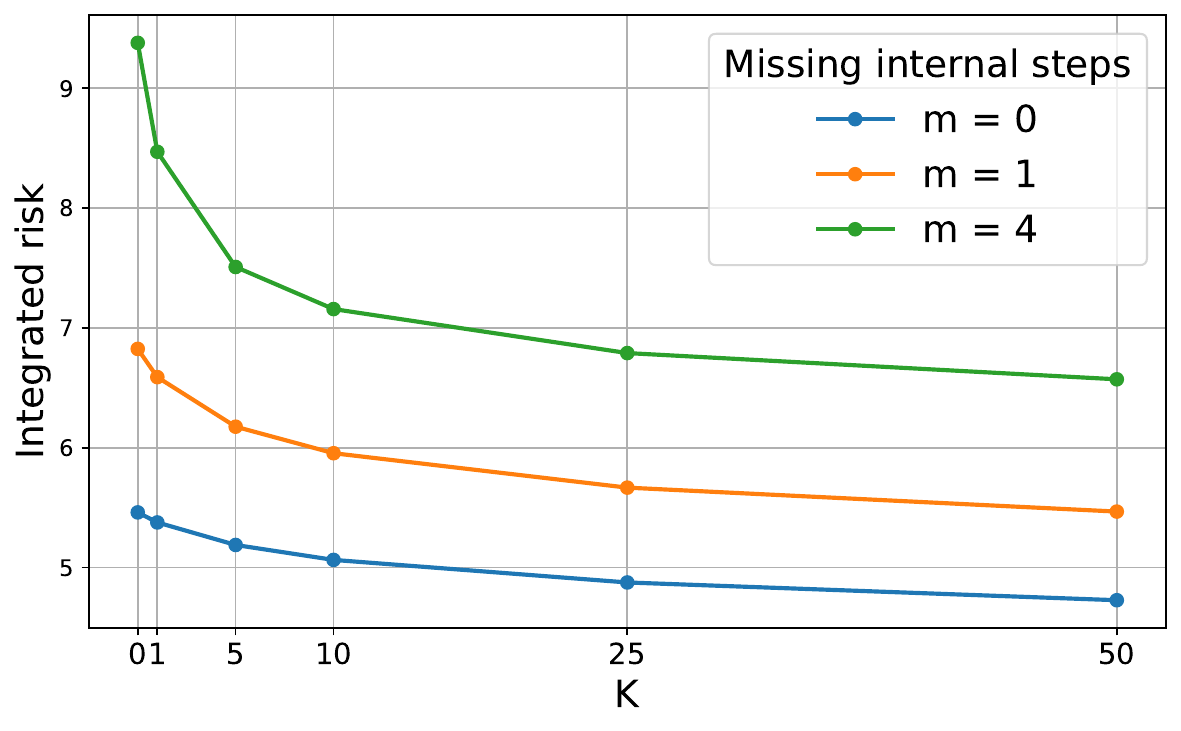}
        \caption{Squared exponential covariance function.}
        \label{fig:risk-vs-K-se-L5}
    \end{subfigure}
    \caption{
    The integrated extrapolation risk $R(D_K, 0, L)$ (lower is better) as a function of the number of external signals $K$, the prediction interval length is $L=5$. 
    The two panels compare the dependence on $K$ for the exponential and squared exponential covariance functions. 
    Each curve corresponds to a different number of missing recent observations $m$ of the target process. 
    }
    \label{fig:risk-vs-K-L5}
\end{figure}

\paragraph{Effect of the number of external series $K$.}
Here, for different values of $m$, we examine how the parameter $K$ affects the integrated squared-error risk.
The results for the two considered covariance functions are in Figure~\ref{fig:risk-vs-K-L5}.
The plots show that, as $K$ increases, the risk converges exponentially to the best possible value given $x_0(t)$ for both covariance functions considered, with a stronger effect for the more local squared covariance function.
So, in the presence of a small amount of noise, it is sufficient to select moderate values of $K$ in the range of $10$ to $50$, depending on the problem.

\subsubsection{Conclusions}

We see that in natural and realistic, albeit still artificial, settings, the gains from using external information are evident and robust, with these gains depending on the availability of recent observations and the number of external series~considered.

%% file: tables/inference_cost.tex
\begin{table}[!ht]
\captionsetup{font=small}
\caption{Additional time and space complexities for different aggregation methods (the vanilla inference and training time is $\mathbf{O(Tm^2)}$, space -- $\mathbf{O(m^2)}$, where $n$ is the number of users, selected for aggregation, $m$ - embedding size, $c$ - hidden size of the MLP for Learnable Kernel attention, and $T$ - is a sequence length.}
\label{tab:inference_cost}
\centering
\resizebox{\columnwidth}{!}{
\begin{tabular}{lcccc}
\hline
 & \multicolumn{2}{c}{\textbf{Inference}}& \multicolumn{2}{c}{\textbf{Training}} \\
\hline
\textbf{Method} & \textbf{Time} & \textbf{Space} & \textbf{Time} & \textbf{Space}\\
\hline
Mean, Max & $O(1)$ & $O(mn)$ & $O(1)$ & $O(mn)$ \\
Exp Hawkes & $O(m)$ & $O(mn)$ & $O(m)$ & $O(mn)$ \\
Attention & $O(mn)$ & $O(mn)$ & $O(mn)$ & $O(mn)$ \\
Learnable Matrix Attention & $O(mn)$ & $O(mn)$ & $O(m^2 + 2mn)$ & $O(mn + m^2)$ \\
Attention Hawkes & $O(mn)$ & $O(mn)$ & $O(mn)$ & $O(mn)$ \\
Learnable Kernel Attention & $O(nm+mc)$ & $O(mn+m^2)$ & $O(m^2 n)$ & $O(mn + m^2)$ \\
Learnable Exp Hawkes  & $O(nm+mc)$ & $O(mn+m^2)$ & $O(m^2 n)$ & $O(mn + m^2)$ \\
\hline
\end{tabular}
}
\end{table}

%% file: chapters_sigmod/04_numerical_experiments.tex



\section{Results}\label{sec:results}

Here, we present the results of experiments on obtaining an external context representation and show how it can be used to enhance existing models.
The subsequent subsections describe the evaluation protocols used.
Then we present the results of the conducted experiments.

\subsection{Evaluation protocols} 
\label{sec:validation_methods}

For evaluating the proposed aggregation methods, we use two experimental setups: we primarily focus on a Classification pipeline and also include additional experiments with a RecSys pipeline. Separate Classification and RecSys pipelines enable the correct evaluation of external aggregation techniques across different machine learning domains, highlighting the efficiency of external context aggregation for varying ML tasks. 

\subsubsection{Classification pipeline}
To validate the proposed methods, we incorporate the benchmark from~\cite{bazarova2024universal}, which considers both local and global embedding properties.

\textbf{Global properties} of representations are validated using the protocol from~\cite{babaev2022coles}.
All datasets considered in this work include a classification label for each sequence. 
For example, this target mark can represent clients who left the bank or failed to repay their loans. The protocol consists of two steps. 
For an initial sequence of transactions of length $T_{i}$ related to the $i$-th user, we obtain the representation $\mathbf{h}^{i} = \mathbf{h}_{i, T_i} \in \mathbb{R}^{m}$, which characterizes the entire sequence of transactions as a whole. 
Given fixed representations $\mathbf{h}^{i}$, we predict the binary target label $y^{i} \in \{0, 1\}$ using gradient boosting.
The hyperparameters of the gradient boosting model are from~\cite{babaev2022coles} and are the same for all base models under study. The quality of a solution to a binary classification problem is measured using standard metrics, like ROC-AUC.

This procedure allows us to evaluate how well the representations capture the user's "global" pattern across its history. 
We call this task \textit{Global target} prediction or \textit{Global validation}.

\textbf{Local properties} of sequences differ from global ones in that they change over time within a sequence. 
We use a sliding window procedure of size $w$ to obtain and evaluate local embeddings. To do this, for the $i$-th user at time $t_{j}\in[t_{w}, T_{i}]$ the subsequence of his transactions $\mathbf{s}_{i, j-w:j}$ is taken.
Next, this interval is passed through the encoder model under consideration to obtain a local representation $\mathbf{h}_{i, t_j} \in \mathbb{R}^{m}$. 




As local validation problems, we explored predicting the next transaction's categorical feature code (for example, MCC code for transactional datasets). We call this task \textit{Event type} prediction. This validation approach was inspired by the work~\cite{zhuzhel2023continuous}, which proposed predicting the type of the next event from the history of observations --— in our case, the categorical feature of the next event. Formally, in this case, the multiclass classification problem is solved: given the internal representation $\mathbf{h}_{i, t_j}$, we predict the categorical code of the event of the $i$-th user, completed at time $t_{j+1}$.

Note that there are many rare MCC types in transactional datasets. From a business perspective, such categories are often less interesting and meaningful. Therefore, in this procedure for testing local properties, it was decided to leave only transactions corresponding to the $100$ most popular codes to simplify the task.

\subsubsection{RecSys pipeline}
\begin{figure}[!ht]
     \centering
     \includegraphics[width=\columnwidth]{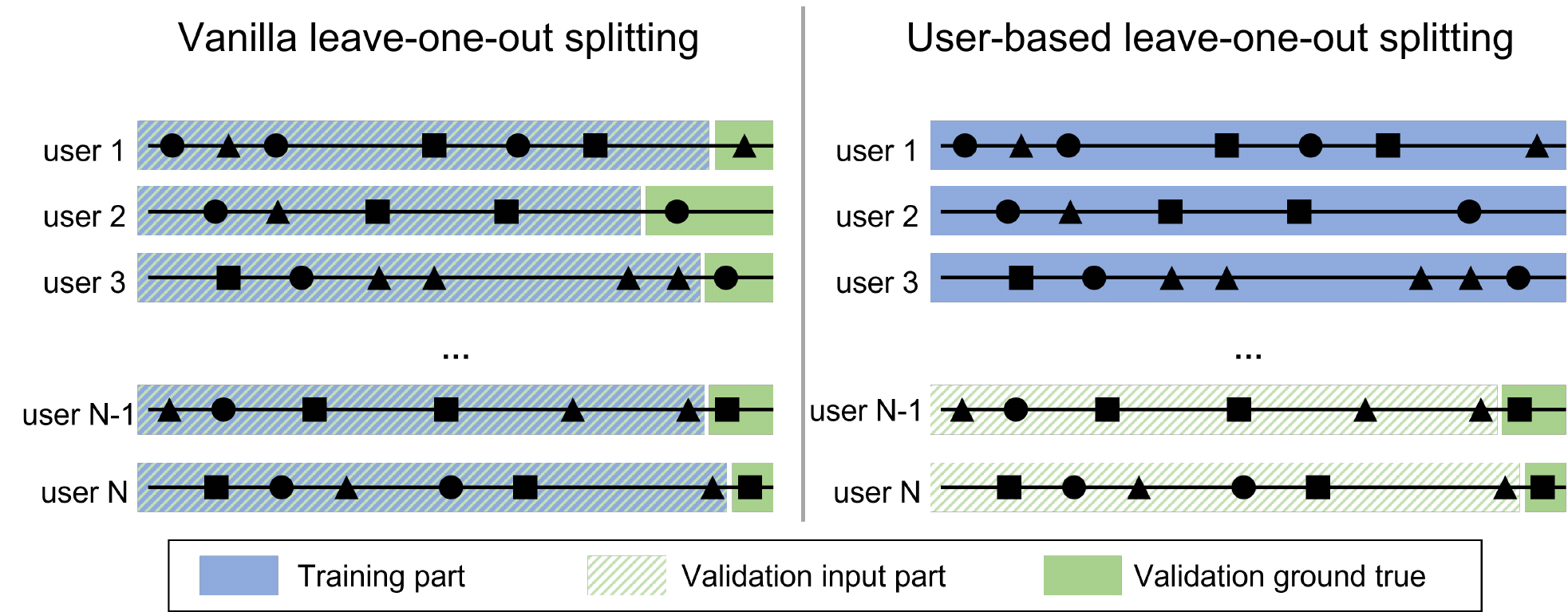}
     \caption{Vanilla leave-one-out split in RecSys (left) vs Our user-based plus leave-one-out splitting (right) to prevent data leaks from external context. }
     \label{fig:recsys_split}
\end{figure}

The procedure is similar to evaluating a recommender system model. We use a combination of user-based and leave-one-out splitting strategies, as shown in Figure~\ref{fig:recsys_split}, to prevent data leakage from external context. User sequences from the training set are used both for training models and external aggregations, and users from the test set are used only for evaluation. 
For the test set, we use all events in the sequence, excluding the last one, as transformer input and compute all metrics for the last event's prediction, as in a vanilla leave-one-out validation procedure. 
However, the test user sequences were not used during model training.
The proposed pipeline can make our results incomparable with other works, using the proposed datasets and models, because the models were not trained on test sequences at all. 
We compare the model with and without an external context using the NDCG@10 ranking metric. 
Unlike local validation, we retain all possible items after dataset cleaning, rather than dropping rare items, to better reflect a real-world scenario.

\subsection{Experimental protocol}
In all experiments, we first pretrain a baseline encoder and compute internal representations using only past events to preserve temporal causality. We then train an aggregation module on top of these representations, either with a frozen backbone or by fine-tuning the aggregation-aware model, and finally evaluate the resulting representations on downstream tasks. For classification datasets, evaluation is performed with global and local validation protocols, while for recommendation datasets, we use the SASRec-based next-item ranking protocol.

\subsection{Methods and Datasets}

We consider three groups of methods: the first without inclusion in the external context, the second with non-learnable aggregation approaches, and the third with learnable aggregations. 
All learnable methods are marked with an asterisk sign~"*".
A conventional encoder~\cite{babaev2022coles} baseline does not use the external information and is labeled \emph{Without context}.
For aggregation approaches, all learnable components of these methods are derived from this self-supervised pipeline, using fixed original encoders. 
We investigated the following types of aggregation of representations to obtain a context vector:
\begin{itemize}
    \item straightforward aggregation methods: \emph{Mean} and \emph{Max} poolings;
    \item methods inspired by Hawkes process: \emph{Exp Hawkes} (Exponential Hawkes), \emph{Learnable exp Hawkes*} (Exponential learnable Hawkes);
    \item attention-based aggregation methods: \emph{Attention} (attention mechanism without learning),  \emph{Learnable attention*} (with learning);
    \item combined method: Hawkes with attention \emph{Attention Hawkes}.
\end{itemize}



To compare the models and methods, we work with nine open samples of event sequences data: \href{https://boosters.pro/championship/rosbank1}{Churn}, \href{https://boosters.pro/championship/alfabattle2}{Default}, and \href{https://www.kaggle.com/datasets/ashisparida/hsbc-ml-hackathon-2023}{HSBC}, as well as six non-financial datasets: 
\href{https://tianchi.aliyun.com/dataset/46}{Taobao} (e-commerce), \href{https://grouplens.org/datasets/movielens/1m/}{MovieLens-1M} and \href{https://grouplens.org/datasets/movielens/20m/}{MovieLens-20M} (cinema), \href{https://cseweb.ucsd.edu/~jmcauley/} {Beauty} (e-commerce), \href{https://remaplab.deib.polimi.it/datasets/}{30Music} (music), \href{https://cseweb.ucsd.edu/~jmcauley/datasets.html}{Beer Advocate} (breweries).

\emph{Churn} dataset contains the transactional data of bank customers and has been used previously~\cite{babaev2022coles}.
For global validation, it is proposed that the problem is whether the client left the bank (churn case).

\emph{Default} dataset~\cite{bazarova2024universal} also contains transactional data of bank customers.
For global validation, it is proposed that the problem is whether the client can repay the loan to the bank.

\emph{HSBC} dataset~\cite{bazarova2024universal} originates from a competition on the fraud identification task.
Each transaction in the dataset is marked to indicate whether the transaction was made by the client personally or maliciously. 
We add the global target, identifying whether at least one transaction in the sequence was fake for each user. In other words, we marked the entire sequence as fraud if any transaction in this sequence was originally marked as fraud. We solve a binary classification task for the whole transaction sequence. In other words, we define the global target as “at least one fraudulent transaction in the sequence,” reflecting a practical risk-detection setting where the goal is to identify users exposed to fraud. 


\emph{Taobao}~\cite{udovichenko2024seqnas} dataset contains logs of user actions from the Taobao platform, including clicks, purchases, adding items to the shopping cart, and favoriting items. 
Initially, the dataset lacked an appropriate global validation task, so we added a global target of predicting whether a customer would make a payment within the next week, following~\cite{udovichenko2024seqnas}. 

\emph{MovieLens-1M}~\cite{babaev2022coles} dataset from the recommender systems domain consists of 1 million explicit ratings on a 5-star scale from 6,040 users on 3,706 movies. 
We use the film genre as a categorical feature and its rating as a numeric feature. Since the dataset doesn't include an explicit global task, we predict users' gender for this task. 
This dataset appears in both the classification modeling and RecSys pipelines.

\emph{Beauty}~\cite{klenitskiy2024does} 
 contains 22K users, 11K items, and 199K events, making it closer to the data from a recommendation system than the financial transactions dataset. It has a stronger sequential pattern than some other RecSys datasets~\cite{klenitskiy2024does}. The dataset contains clothes and accessories purchased from Amazon. We use it only for the RecSys pipeline validation.

 \emph{Beer Advocate}~\cite{mcauley2013hidden}
 is a large collection of user-generated beer
 reviews sourced from the Beer Advocate website, comprising about 1.5 million reviews from 33K users with
 ratings on multiple aspects. The sequence of data is user reviews. The problem of predicting the beer style of the next review is solved.

\emph{30Music}~\cite{turrin2015playlists}
 contains 45K users, 5.6M tracks,
 50K playlists, and over 31M play events. In addition, it includes explicit
 user preferences in the form of loved tracks and user-generated playlists.
 We use each user's listening events as sequential signals. The dataset is used for RecSys pipelines.

\emph{MovieLens-20M}~\cite{harper2015movielens}
 from the MovieLens
 movie recommendation service, containing 20,000,263 explicit ratings across 27,278 movies by 138,493
 users. It is similar to the MovieLens-1M. Only the RecSys pipeline was used on the dataset.

Table~\ref{tab:datasets} summarizes the main characteristics of all presented datasets.

\input{tables/dataset}

\subsection{Implementation details}
\label{subsec:implementation_details}

This section summarizes the exact experimental setup used throughout the paper. We first specify the backbone architectures and input features for classification, then describe the training settings of the aggregation modules, present the downstream evaluation heads for the classification pipeline, and finally describe the RecSys pipelines. Unless stated otherwise, the main backbone hyperparameters are given in Table~\ref{tab:hyperparams_trans}.

\paragraph{Backbone.} 

In this research, the pipeline follows that of the \href{https://github.com/dllllb/pytorch-lifestream/tree/main}{pytorch-lifestream} package, which contains an implementation of the CoLES model. 

We use an RNN-based or Transformer-based architecture as the model backbone.
Depending on the dataset, the RNN block is either a Long Short-Term Memory (LSTM) or a Gated Recurrent Unit (GRU) in an RNN-based backbone~\cite{babaev2022coles}. 
We kept the number of trainable parameters consistent across all encoder architectures, excluding RecSys validation.
For all datasets in the classification pipeline, the model takes two features as input: numerical and categorical, excluding the Taobao dataset, which contains only a categorical feature. We treat the MCC code for Churn, Default, and HSBC, the item ID for Taobao and MovieLens. For numerical features, we take amounts of transactions for Churn, Default, HSBC, and ratings for the MovieLens dataset. 
For categorical variables, we use embedding of size $d_{\mathrm{mcc}} = 24$ for Churn and HSBC or $16$ for Default. Numerical features, scaled with the standard scaling method, concatenated with categorical embeddings, become the input to the RNN or Transformer encoder. 
For the RNN model, we set the number of recurrent layers equal to one. 
In experiments with the Transformer architecture, we used a configuration with two layers and two attention heads per layer. The MCC variable is embedded using a fixed size of $d_{\mathrm{mcc}} = 24$. The rest of the hyperparameters are listed in Table~\ref{tab:hyperparams_trans}.


Consistent with the methodology, the baseline encoder is pretrained in a self-supervised manner using the same objective as in the original pipeline: a contrastive CoLES loss or a generative next-event-type prediction loss for non-RecSys datasets, and a cross-entropy next-item prediction loss for RecSys datasets.

\input{tables/hyperparams}

\paragraph{Aggregation parameters.}
For Hawkes-based parameters, we tune the $\ gamma$ parameter, setting it close to $1e-11$ due to the large timestamp values.

The learnable attention matrix $A$ from the matrix attention method and the learnable function $\phi(\cdot)$ from the kernel attention and exponential learnable Hawkes methods were trained in the same learning pipelines with a fixed encoder. 
For the learnable functions $\phi(\cdot)$, we use a two-layer neural network with a hidden size equal to $100$. 
All the learnable elements were trained for $60$ epochs with a batch size of $128$.

A model using external information needs to store internal representations of all users from the training set for all time points.
Due to memory constraints, we use only a random subset of internal representations. For the Churn dataset, the number of clients to train on in all experiments was $1000$; for all other datasets, $300$ for the classification pipeline. For the RecSys pipeline, we use $100$ users for external context aggregation due to the higher computational demands of RecSys models. 
As the ablation study shows, it is sufficient to represent the external context.

All learnable aggregation methods are trained within the same protocol on top of pretrained internal representations. In the main setup, the encoder is kept fixed during aggregation training unless fine-tuning is explicitly stated. The number of sampled external users is restricted by memory and computational constraints and is therefore selected separately for the classification and RecSys pipelines.

\paragraph{Downstream heads.} 

For the global validation task, we train the LightGBM model~\cite{ke2017lightgbm} with $500$ estimators, a learning rate equal to $0.02$, and regularization coefficients for $l_1$ and $l_2$ regularization equal to $1$. For the local validation task, we train the linear head over the encoder. There, the hyperparameters are the following: window size equal to $32$, stride equal to $16$, batch size equal to $512$, and a maximum number of epochs equal to $20$ for Churn, Default, Taobao, and MovieLens datasets, and $10$ for HSBC, learning rate equals $0.001$, optimizer is Adam.

All results were averaged across $3$ training runs on the Nvidia L40 GPU for pre-trained encoder models.


\paragraph{RecSys pipeline.} For filtering data, we drop all users and items with fewer than five interactions. 
Besides, we saved in the datasets only interactions with positive feedback, where the rating was higher than three. We set the maximum sequence length to 64 for all datasets, except for the MovieLens 1M dataset, which is set to 200. Models were trained and fine-tuned with the Adam optimizer and a learning rate of 0.0001.
We train the aggregation model for 10 epochs across all RecSys datasets and for 20 epochs on the MovieLens 1M dataset. We take the full cross-entropy loss.
The model includes 2 transformer blocks with 2 attention heads. We set the weight coefficient $\alpha$ to 0.01 or 0.1 for different datasets, empirically determining this value. 

Unless stated otherwise, all RecSys experiments follow the same preprocessing and optimization protocol: filtering users and items with fewer than 5 interactions, keeping only positive feedback events, using a fixed maximum sequence length, training with Adam at a learning rate of 0.0001, and optimizing the model with the full cross-entropy loss. The remaining dataset-specific settings, including the number of epochs and batch size, are as reported in this subsection and in Table~\ref{tab:hyperparams_trans}.


\subsection{Main results}

\input{tables/ranks_4}

\begin{figure}[!ht]
     \centering
     \includegraphics[width=\columnwidth]{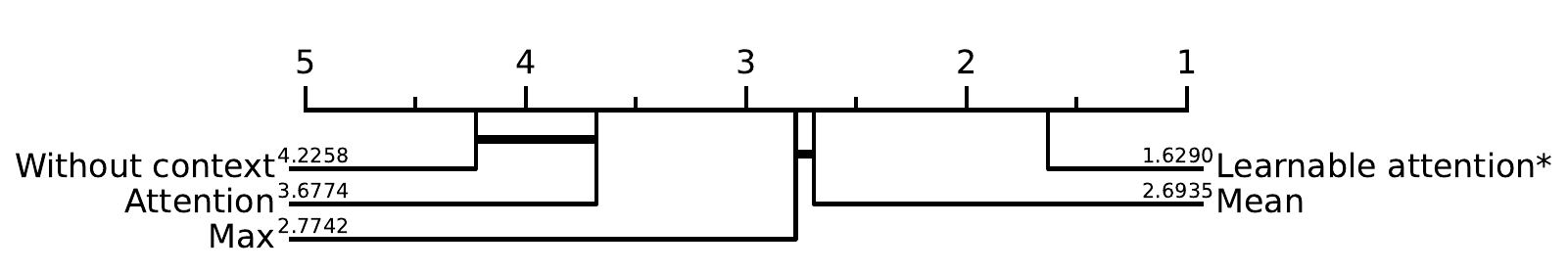}
     \caption{CD diagram with statistical significance for Wilcoxon-Holm post-hoc analysis for all aggregation methods, presented in RecSys and Classification pipelines, including experiments with generative loss and transformer backbone. The Learnable attention method is statistically significantly better than other aggregations with $\alpha$ =0.05. }
     \label{fig:cd}
\end{figure}


The main results, aggregating all experiments with both the RecSys and the Classification pipeline, are presented in the Critical difference (CD) Diagram \cite{fawaz2019deep} in Figure~\ref{fig:cd}. This diagram describes the statistical significance of the ranks of the proposed methods. For building this plot, we include experiments, presented later in Tables~\ref{tab:ranks}, ~\ref{tab:metrics_table_transformer}, and ~\ref{tab:generative_metrics}, and Figure~\ref{fig:recsys}. CD Diagram doesn't contain Hawkes-based aggregation methods because we didn't include them in the RecSys pipeline experiments due to the need for computationally intensive hyperparameter $\gamma$ tuning. As we can see, the Learnable Attention method outperforms all other aggregation methods and the baseline without aggregation. On the other hand, Mean and Max aggregation yield almost equivalent results due to the similarity of the ideas. The aggregated ranks for the Attention method without learnable parts outperform the vanilla model without aggregation, but the difference is not significant. Additional details about statistical significance are presented in Section~\ref{sec:cd} and Table~\ref{tab:cd}. 

\subsubsection{Classification pipeline validation}
The full results obtained with an RNN encoder with contrastive loss function, trained in a contrastive manner, are presented in Table~\ref{tab:ranks}, while
Figure \ref{fig:mean_rank} summarizes them by presenting the average ranks of the ROC-AUC metric over all classification datasets and methods. We also present Table~\ref{tab:pr-auc} with the PR-AUC metric for financial datasets with class imbalance. In additional materials, we also provide Table~\ref {tab:metrics} with extended results from Table~\ref{tab:ranks}.



\input{tables/prauc}

Accounting for external context typically improves metrics. 
This is especially noticeable in the class-balanced Churn dataset. In the unbalanced Default and HSBC datasets, contextual representations also help models solve both local and global problems, as measured by ROC-AUC and PR-AUC metrics; however, this effect is less pronounced. 
The benefits of external context appear consistently across domains, albeit with varying impact. 
The Taobao and MovieLens 1M datasets also demonstrate improvements in both local and global scenarios.

According to Figure~\ref{fig:mean_rank}, the best method is the Learnable attention, which performs best or second best for all but one task.
 The Learnable attention method is the best or second-best for all presented cases in the global validation scenario, demonstrating the importance of attention mechanisms for solving tasks, including entire sequences. 
It is natural, as attention-based probing often yields stronger models by considering more intricate user connections~\cite{velivckovic2018graph}.  However, for some cases, e.g., Default, simpler attention aggregation may outperform Learnable attention. This could be due to the dataset's complex structure or to class imbalance. On the other hand, as shown in Table~\ref{tab:pr-auc}, Learnable attention demonstrates better results than attention without learnable parameters, according to PR-AUC metrics, which is more sensitive to class imbalance.
On the other hand, the Learnable exp Hawkes method is efficient for local tasks, especially on financial datasets, because it is designed with a temporal component that can capture local patterns. Approaches inspired by the Hawkes process could perform well but require finding the $\gamma$ parameter. Learnable attention is a more preferable method than Hawkes-based methods.
Thus, one can easily include the external context without modifying the overall architecture.
However, building a model with a learnable external context aggregation method, like \emph{Learnable attention}, provides further improvements.

\subsection{Additional results}

\begin{figure*}[!ht]
\centering
\begin{subfigure}{.6\columnwidth}
  \centering
  \includegraphics[width=.97\linewidth]{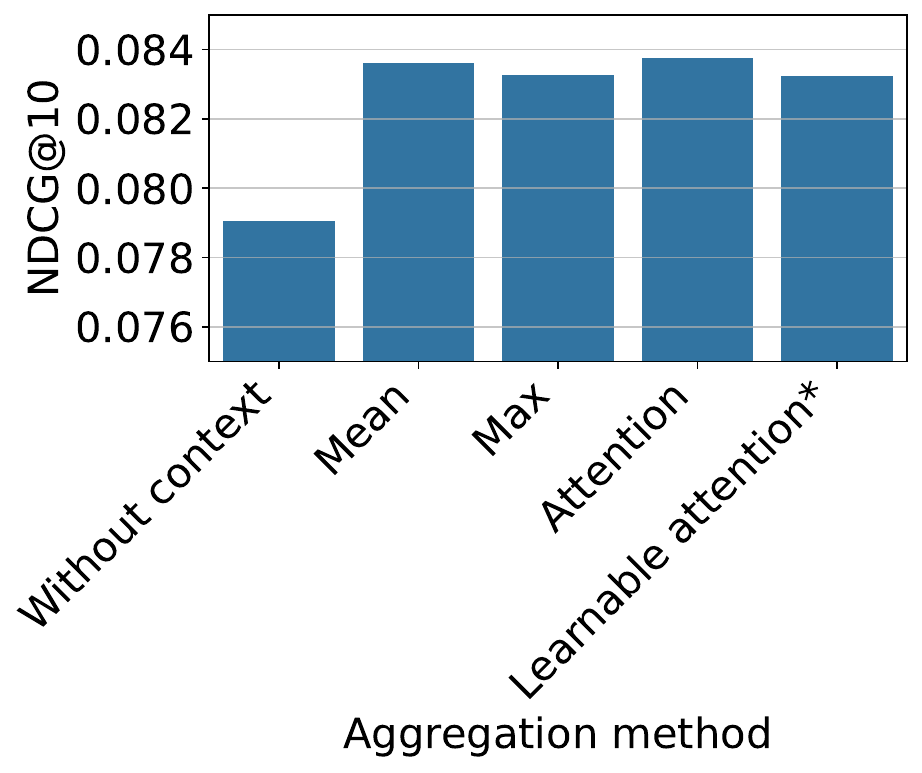}
  \caption{MovieLens 1M}
  \label{fig:recsys_movielens}
\end{subfigure}
\begin{subfigure}{.6\columnwidth}
  \centering
  \includegraphics[width=.97\linewidth]{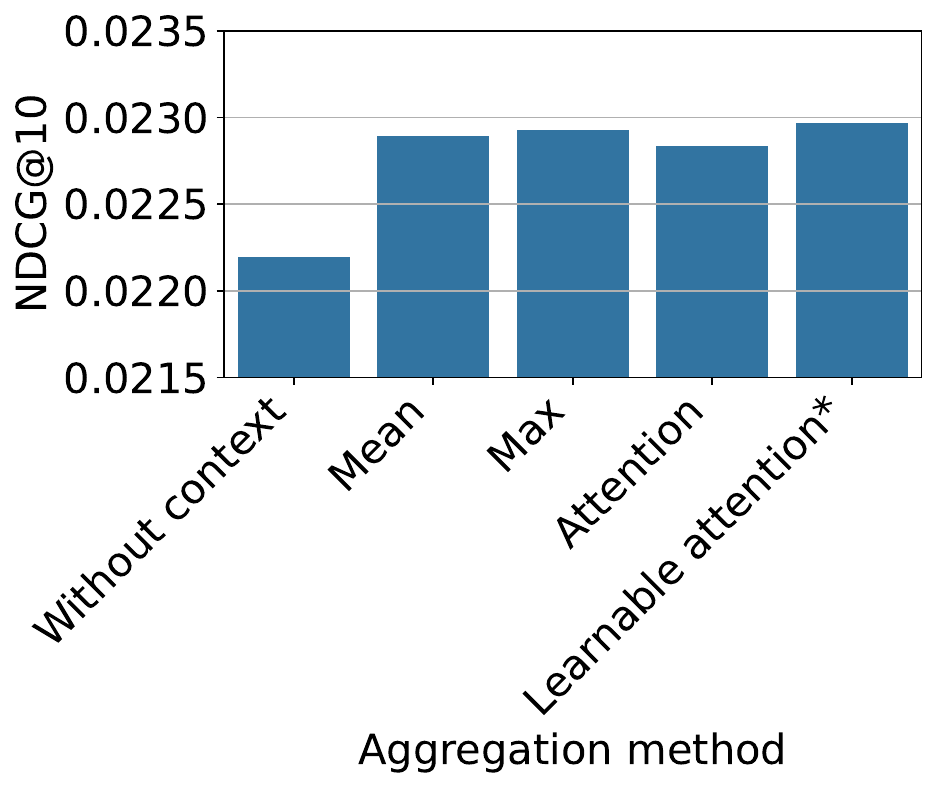}
  \caption{Beauty}
  \label{fig:recsys_beauty}
\end{subfigure}
\begin{subfigure}{.6\columnwidth}
  \centering
  \includegraphics[width=.97\linewidth]{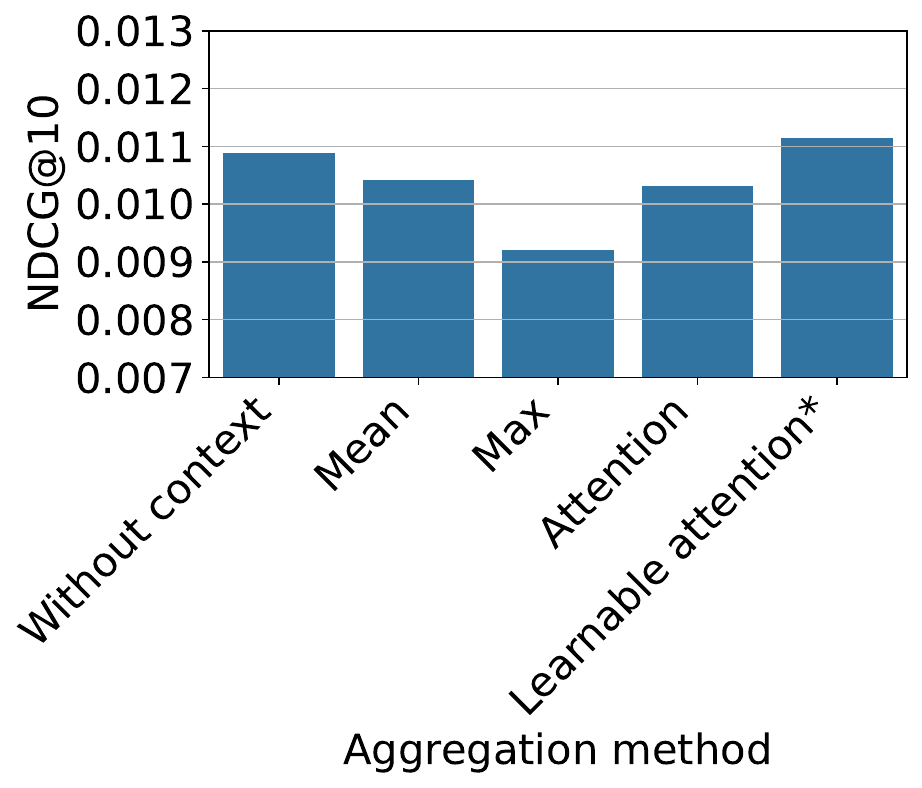}
  \caption{Beer Advocate}
  \label{fig:recsys_beer_advocate}
\end{subfigure}
\begin{subfigure}{.6\columnwidth}
  \centering
  \includegraphics[width=.97\linewidth]{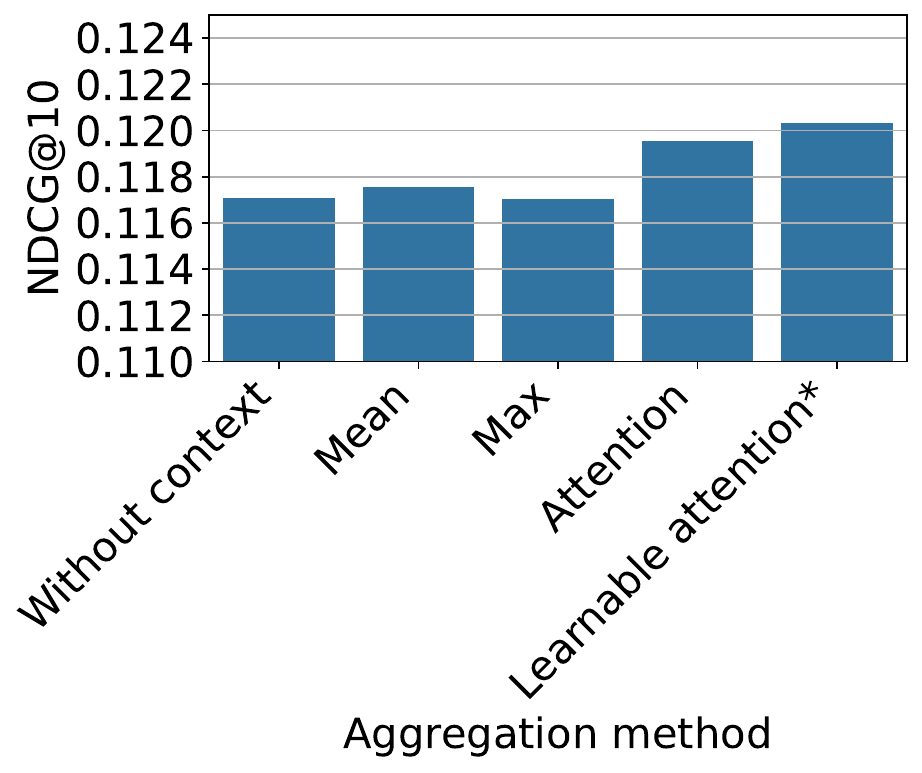}
  \caption{30 Music}
  \label{fig:recsys_30music}
\end{subfigure}
\begin{subfigure}{.6\columnwidth}
  \centering
  \includegraphics[width=.97\linewidth]{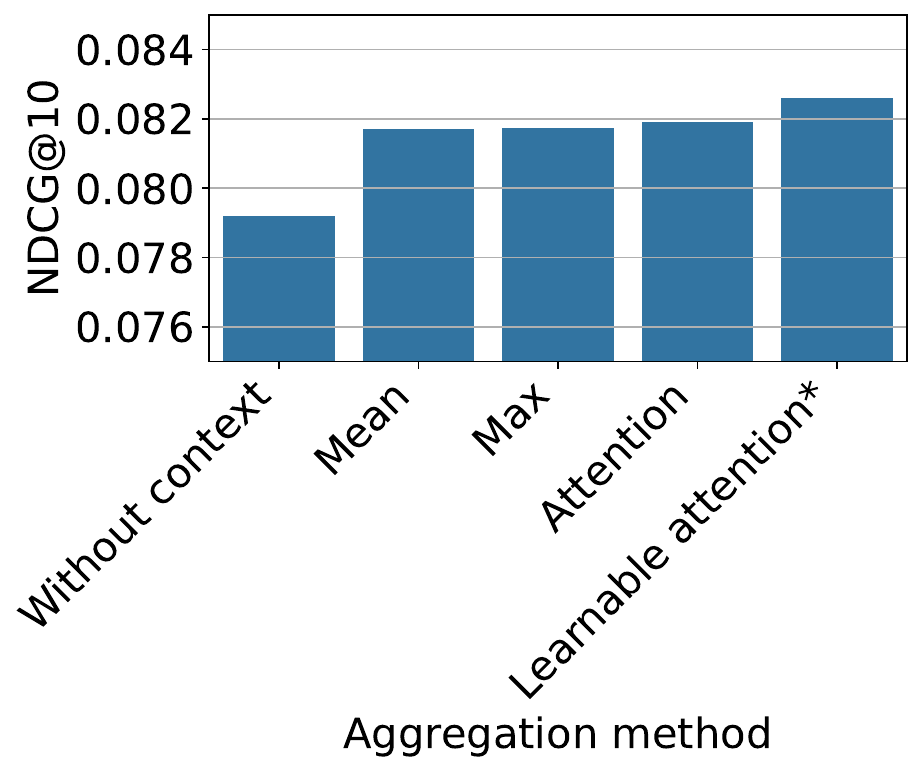}
  \caption{MovieLens 20M}
  \label{fig:recsys_movielens20m}
\end{subfigure}
\caption{NDCG@10 metrics for SASRec model with and without external aggregation for all RecSys datasets.}
\label{fig:recsys}
\end{figure*}

In this part of the work, we provide additional experiments to further validate external context aggregation. We include sanity checks, evaluation in alternative setups (RecSys, transformer encoder, and generative training), and a sensitivity analysis of key hyperparameters. We also assess robustness to adversarial perturbations and missing events, and analyze the computational overhead of the proposed approach.

\subsubsection{Sanity Checks for External Context Aggregation}

\paragraph{External aggregation for baseline predictions}

In this experiment, we want to demonstrate the effectiveness of external context aggregation in much simpler cases. In this experiment, we predict the next transaction's MCC by using the most popular MCC code. In an internal case, we take the most popular transaction from a user sequence. In the internal + external method, we also use the most popular MCC code from a set of user transactions, extended with the last transactions from external sequences.
As shown in Table~\ref{tab: baseline}, the external context boosts the baseline method's prediction quality for two of 
three datasets. This basic experiment proves the efficiency of external information aggregation for sequential data.

\input{tables/external_baselines}

\paragraph{Predictions based only on external context.}

In this experiment, we validate the ability of models built solely on external information features to perform effectively across different tasks. To conduct the experiment, we change the internal context using a vector of random values. 

Table \ref{tab:only_external} compares the performance of models built using only external aggregation embeddings across two pipelines: a recommendation system on the MovieLens-1M dataset and a sequential model on a Churn prediction dataset. For the RecSys task, attention-based aggregation substantially improves NDCG@10 over simple mean pooling, with learnable attention achieving the best result. A similar trend is observed in the sequential pipeline, where attention mechanisms outperform mean aggregation on both global and local validation splits, yielding higher overall ROC-AUC scores. So, as shown, the external information contains useful information for applied problems.

\input{tables/only_ext}

\subsubsection{Additional validation scenarios}
In this subsection, we consider a different type of external context aggregation for sequential data scenarios: a recommendation problem, a transformer architecture instead of an RNN, and a generative next-item prediction pretrain loss instead of contrastive loss in the main pipeline.


\paragraph{RecSys pipeline validation}

For the RecSys pipeline, the experimental results appear in Figure~\ref{fig:recsys}. 
We see a boost in metrics for all datasets when compared with results without external context. Because of the higher computational complexity compared with the classification pipeline and the need to estimate the $\gamma$-coefficient for Hawkes-based aggregations, we didn't include this aggregation in these experiments. 
Across almost all RecSys datasets, excluding MovieLens 1M, we observe the dominance of the Learnable attention method. The MovieLens 1M case can be explained by the dataset's simplicity, which allows basic aggregations to demonstrate better performance. In a larger MovieLens-20M with a close structure, we can see the dominance of the learnable attention method. Additionally, across all datasets, external aggregation outperforms the vanilla model. 

\begin{figure*}[!th]
     \centering
     \includegraphics[width=\textwidth]{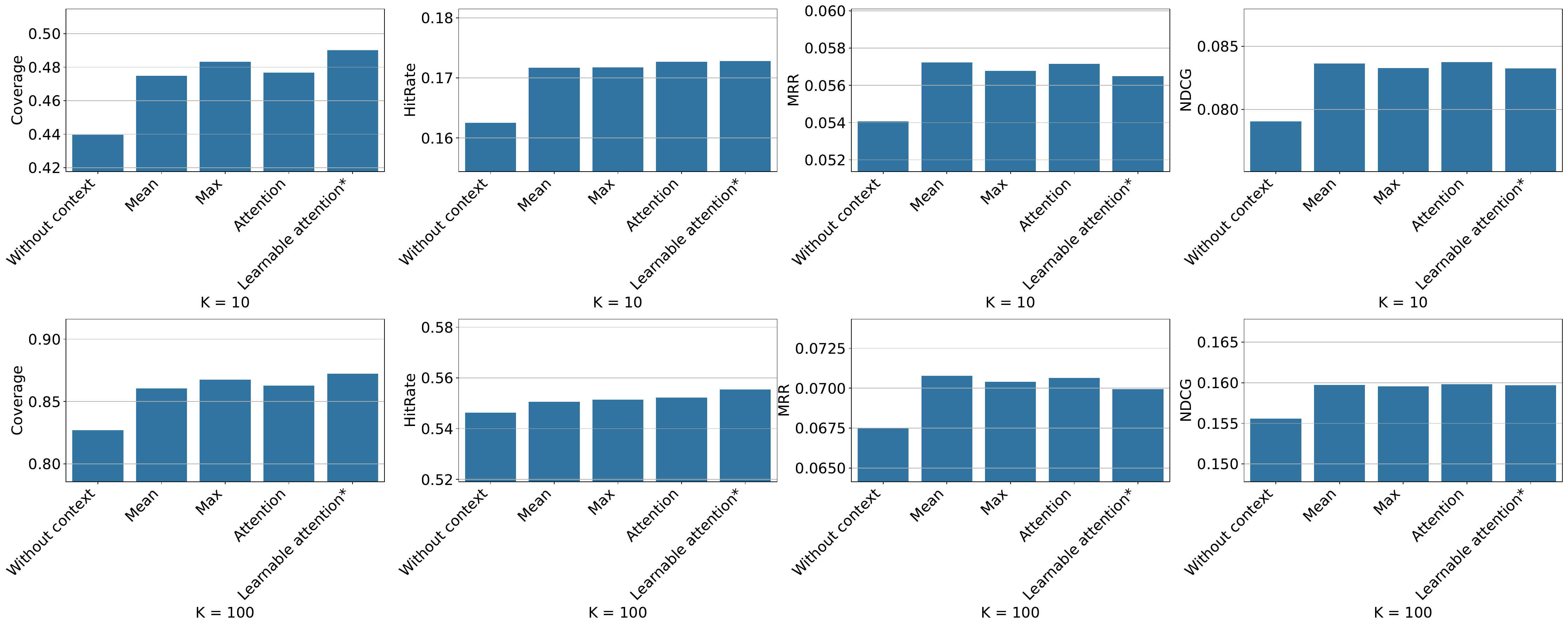}
     \caption{Different metrics for the RecSys pipeline and MovieLens 1M. All metrics are calculated for 10 and 100 recommended objects.}
     \label{fig:recsys_ml_multi}
\end{figure*}

Besides, Learnable attention demonstrates good performance not only for NDCG@10 metrics, as shown in Figure~\ref{fig:recsys_ml_multi}. We can see that Learnable attention shows significant improvements in HitRate and Coverage metrics, despite not performing well on NDCG@10 metrics for this dataset. Moreover, metrics increase for both 10 and 100 recommended items because of the external aggregation technique.

As noted in Section \ref{sec:validation_methods}, the presented metrics may not be comparable to those in other works because many studies use only a leave-one-out pipeline. In our work, we combine it with user-based splitting to avoid data leaks from external context, which can make our metrics lower compared to the vanilla leave-one-out pipeline. For instance, we obtain 0.1247 NDCG@10 on the classic leave-one-out split and 0.0776 with our split method on the MovieLens-1M dataset for the SASRec model without external aggregation. Our results are comparable to another work that tested a different splitting strategy~\cite{gusak2025time}.

\input{tables/ar}

\paragraph{Transformer instead of RNN}

We conduct an additional experiment by replacing the RNN encoder with a transformer-based encoder.
Table~\ref{tab:metrics_table_transformer} reports the results.

\input{tables/transfromers_2}

The evidence confirms the overall trends observed. The best performance for this architecture appears when context aggregation methods are used. Besides, Learnable attention remains the best method for external context aggregation. The transformer architecture without external context may not be as efficient as RNN for some tasks, for example, Churn and Taobao local validation, and HSBC and MovieLens for global validation, but external context methods help to increase metrics almost to the level of the RNN model for these cases. It demonstrates that external aggregation methods can be highly efficient when the encoder model is not optimal.

\paragraph{Generative instead of Contrastive}

In this experiment, we pre-trained and then trained a learnable aggregation method in a generative manner for next-type transaction prediction, rather than using contrastive loss. The results are presented in Table~\ref{tab:generative_metrics}. As we can see, external aggregation also improves model performance on both local and global validation tasks, demonstrating the dominance of the Learnable attention method, especially after fine-tuning. The exclusion is the HSBC local validation case, where strong baseline-model performance yields better results, and basic aggregation methods perform better.

\subsubsection{Sensitivity study}


In the sensitivity study, we consider how major hyperparameters of a method, the number of external sequences, the temperature parameter $\gamma$ for Hawkes-based methods, the external weight parameter $\alpha$ for the RecSys scenario, and the possibility of fine-tuning the backbone, affect the quality.

\paragraph{Dependence on the number of external sequences}

We measured how context size affects aggregation quality by running tasks with the number $n$ of auxiliary sequences ranging from $10$ to $1000$. The evaluation uniformly randomly selects users whose last events occurred in the most recent time step. 
The considered methods are Mean, Learnable Attention, and Exp Hawkes.

\begin{figure}[!t]
     \centering
     \includegraphics[width=0.99\columnwidth]{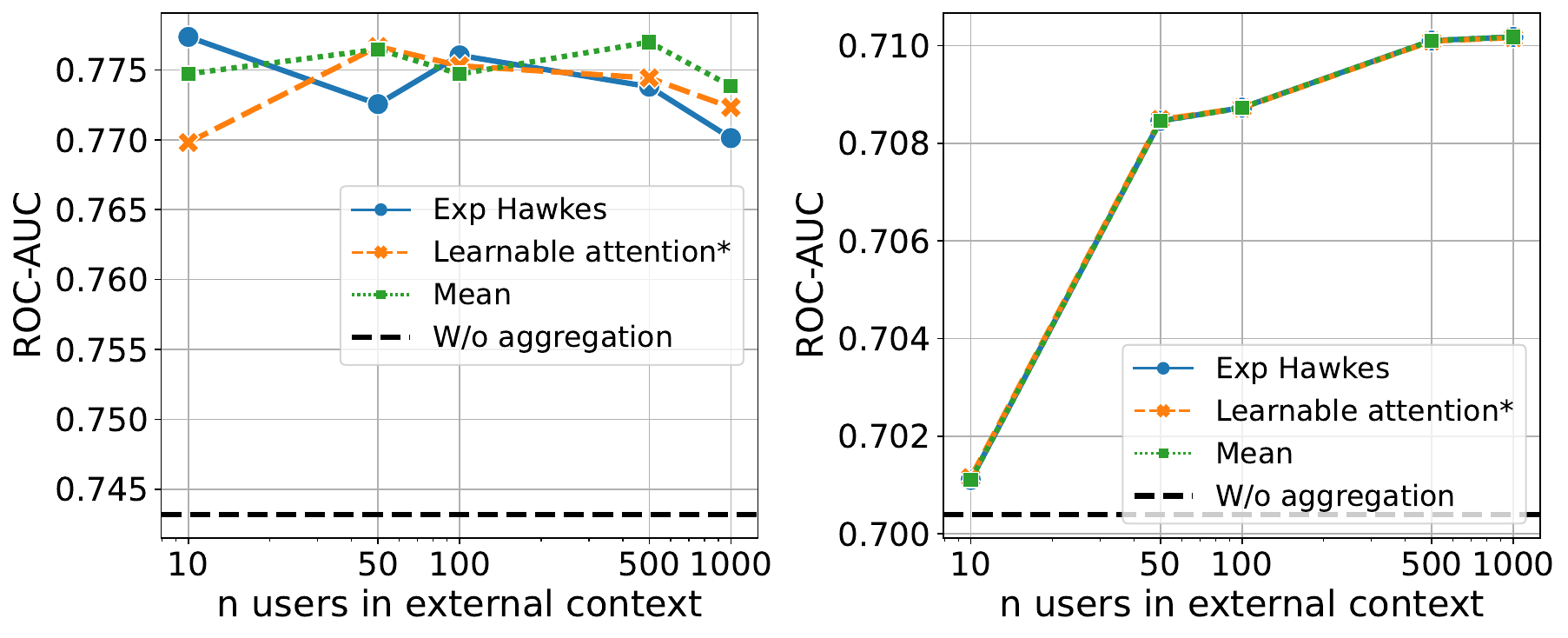}
     \caption{Dependencies between the number of event sequences in the external context for the global (left) and local (right) targets and ROC-AUC values for \textit{Churn} dataset through one run. }
     \label{fig:iters}
\end{figure}

Figure \ref{fig:iters} demonstrates the results for the Churn dataset through one run. Increasing the number of selected users for aggregation
steadily boosts performance on both tasks for Mean and Learnable Attention. Marginal returns flatten once the context window reaches about 500 sequences, implying this threshold is sufficient to attain top accuracy. 
In comparison, the Exp Hawkes model doesn't show an improvement on the global validation task, suggesting limited capacity to exploit cross-user signals, but it does show a significant improvement on the local validation task. 

\paragraph{Dependence on the Hawkes decay parameter}

\begin{figure}[!t]
    \centering
    \includegraphics[width=0.92\columnwidth]{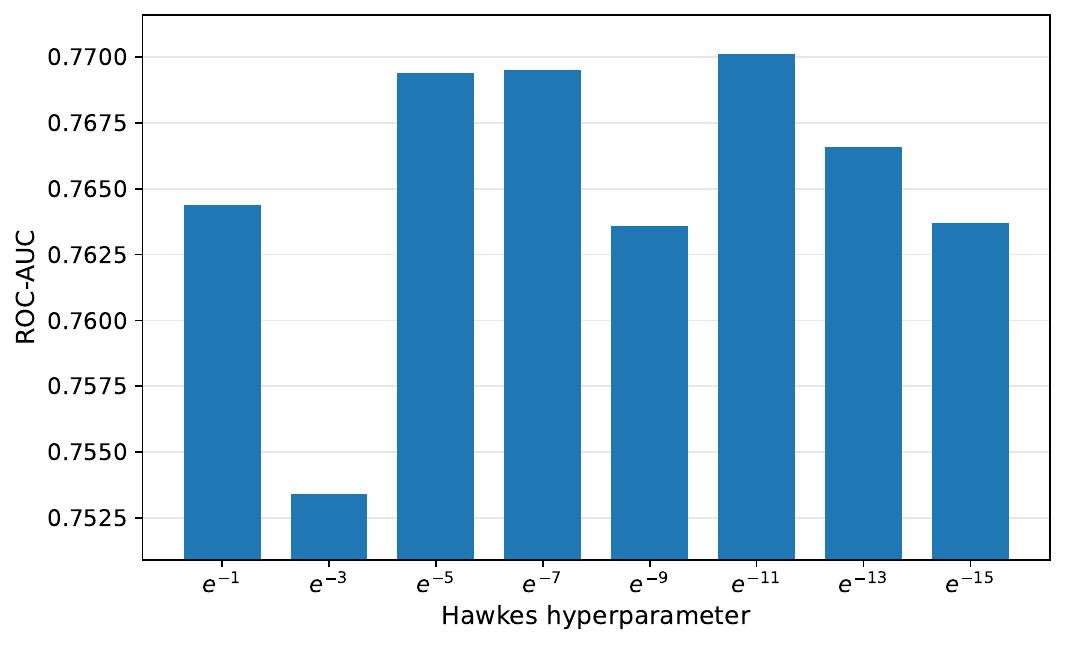}
    \caption{ROC-AUC as a function of the Hawkes decay hyperparameter $\gamma$ for the Exp Hawkes aggregation on the Churn dataset under global validation.}
    \label{fig:hawkes_gamma}
\end{figure}

As performance of Hawkes-based methods directly depends on the coefficient $\gamma$, one needs to find a reasonable $\gamma$ for the method to perform well. Figure~\ref{fig:hawkes_gamma} depicts ROC-AUC as a function of $\gamma$ for Churn dataset under global validation. Best result is achieved for $\gamma_0 = e^{-11}$. The decay hyperparameter $\gamma$ is tuned empirically for other datasets while taking $\gamma_0$ as a reference point.

\paragraph{Dependence on the external weight parameter for Recsys}

\begin{figure}[!ht]
     \centering
     \includegraphics[width=\columnwidth]{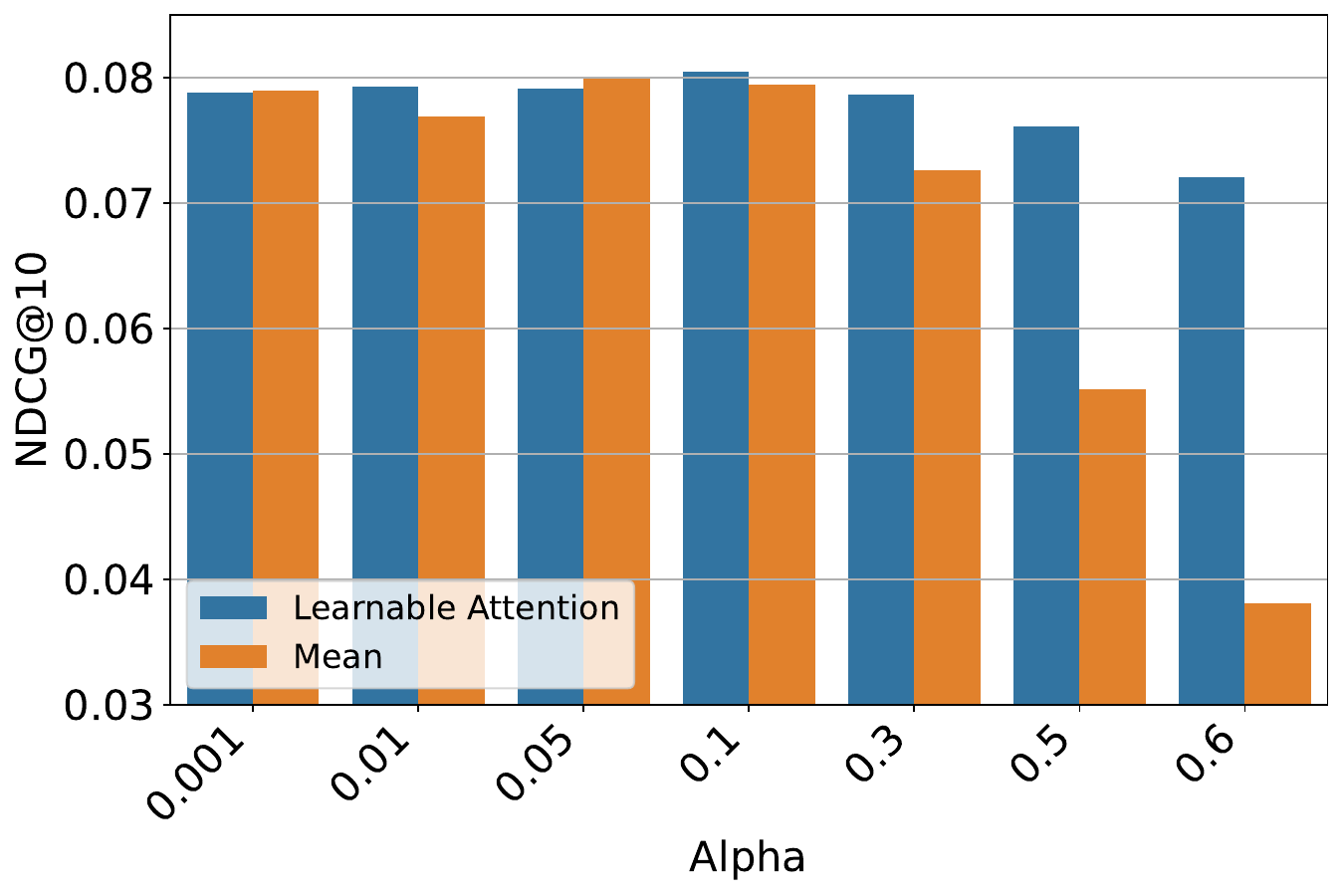}
     \caption{NDCG@10 for RecSys pipeline and MovieLens 1M dataset for different $\alpha$ (alpha) values - the share of external context in final user embeddings.}
     \label{fig:recsys_alpha}
\end{figure}

We determine the optimal $\alpha$ parameter for the share of external context in the RecSys pipeline. The results are presented in Figure~\ref{fig:recsys_alpha}. The Learnable attention method performs better when $\alpha$ equals 0.1, but mean aggregation is better with $\alpha=0.01$. It also shows that Learnable attention external representations can effectively be combined with internal aggregation vectors, allowing us to leverage a stronger signal from the external context. Besides, if we set the external information sharing parameter $\alpha$ extremely large, 0.5 or higher, we see a strong decrease in quality for the basic Mean aggregation method, but the Learnable attention mechanism helps maintain metrics at a high level.
The Learnable attention method outperforms the Mean aggregation across almost all values of $\alpha$, especially for large values.

\paragraph{Fine-tuning backbone for local validation}

To improve the quality of our aggregation methods, we unfreeze the backbone and the model's event-type prediction head in the classification pipeline. 
This procedure is only applicable for local validation, as using a differentiable linear head instead of gradient boosting significantly decreases the quality of the global validation task. 
We compare training the neural network in two modes: training both the encoder and the head (unfrozen mode) and training only the head with a frozen internal representation encoder, as in all previous experiments. 

\begin{figure}[!h]
\centering
\includegraphics[width=.9\linewidth]{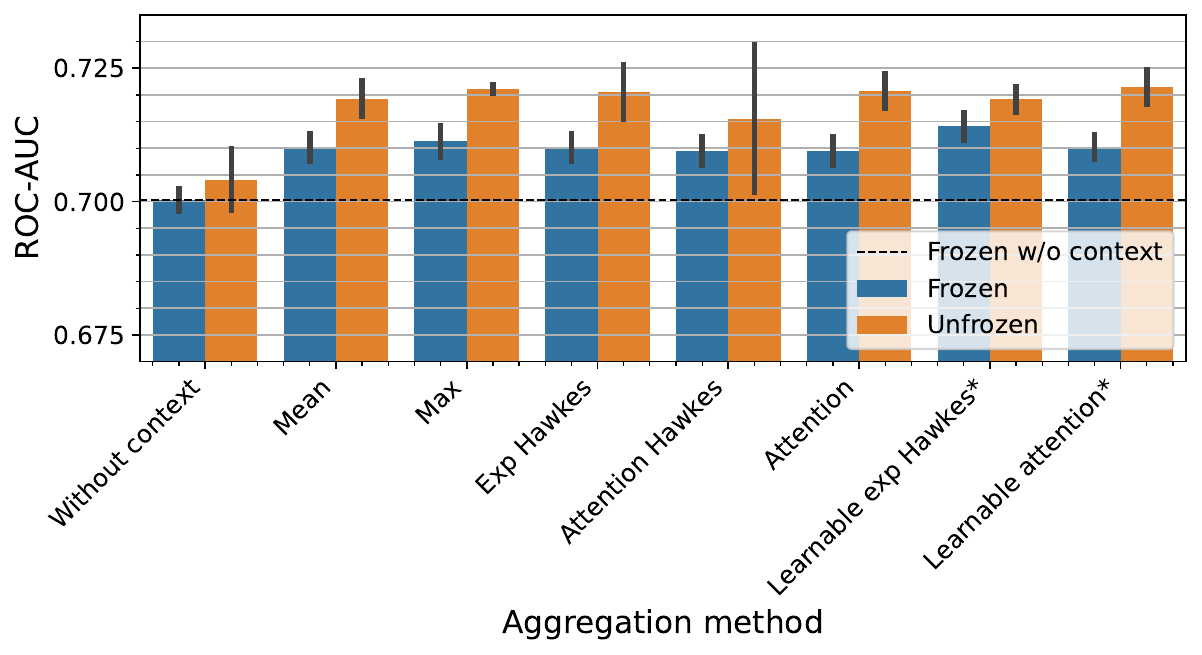}
\caption{ROC-AUC for the Churn dataset and task next event type prediction without (Unfrozen) and with (Frozen) fine-tuning the encoder.}
\label{fig:fine-tune}
\end{figure}

The results are presented in Figure \ref{fig:fine-tune}. 
Unfreezing the encoder improves aggregation quality, especially for the Churn dataset. 
Additionally, the ROC-AUC for all aggregations increased substantially compared to the encoder without external context.  
Learnable attention remains the best approach among those considered. Moreover, unfreezing the encoder improves the quality of external aggregation methods much more than a model without the external context.

\subsubsection{External aggregation robustness}

In this section, we present and discuss experiments demonstrating the robustness of external aggregation to adversarial perturbations and to sequences without relevant events.

\paragraph{Adversarial robustness}

\begin{figure}[!h]
\centering
\includegraphics[width=.9\linewidth]{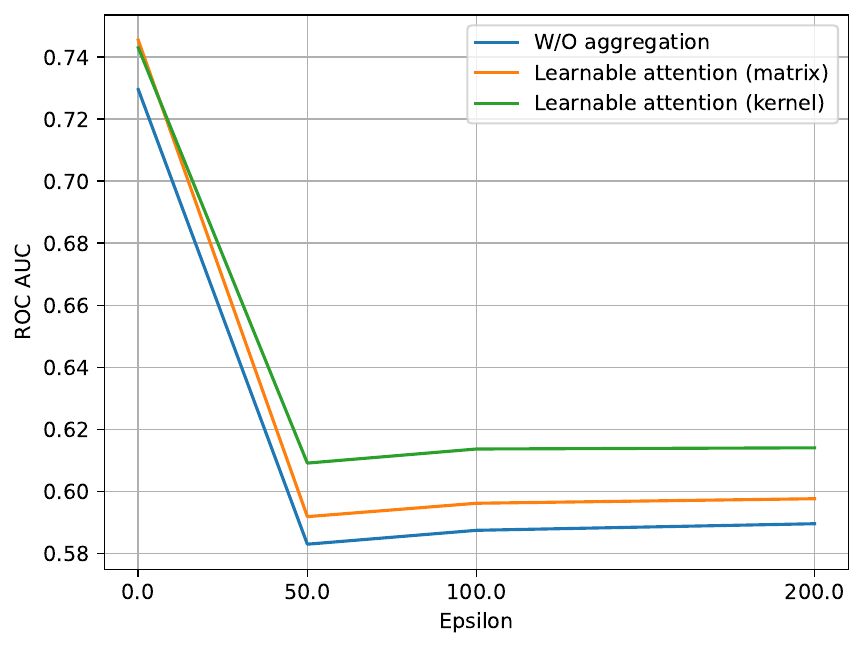}
\captionsetup{font=small}
\caption{Decreasing ROC-AUC metric under concat FGSM attack with different attack strength (epsilon) for CoLES model, Churn dataset, global validation pipeline. Epsilon 0 is equal to the case without an attack.}
\label{fig:attack_notrand}
\end{figure}

External context aggregation should have one additional important feature: it could increase model robustness to adversarial perturbations. The reason is that adversarial perturbations could be applied to a single sequence, so the quality of the internal vector decreases while the external features remain unchanged. 
One of the adversarial attack methods for transaction data is Concat FGSM~\cite{fursov2021adversarial}, adding specially generated perturbed transactions to the end of the sequence. The idea of the attack is to add transactions and then change them to more decreasing model quality, finding the closest transaction in the embedding space to the initially perturbed transaction .
In this experiment, we attacked the CoLES model on the Churn dataset for global validation scenario, with and without external aggregation, at different attack strengths $\epsilon$.  However, unlike the general global validation pipeline, we use an encoder-head neural network rather than boosting over embeddings because we need gradients to provide attack. The results, presented in Figure~\ref{fig:attack_notrand}, demonstrate that Learnable attention, especially kernel attention, dropped the metric less compared with the method without aggregation. The presented results are aggregated by runs with different seeds. The increase in performance for large epsilon could be explained by the saturation of the attack, while for smaller than 50 $\epsilon$, the initially generated attack doesn't change the transaction MCC, so this attack is equivalent to a random one.

\begin{figure}[!h]
\centering
\includegraphics[width=.9\linewidth]{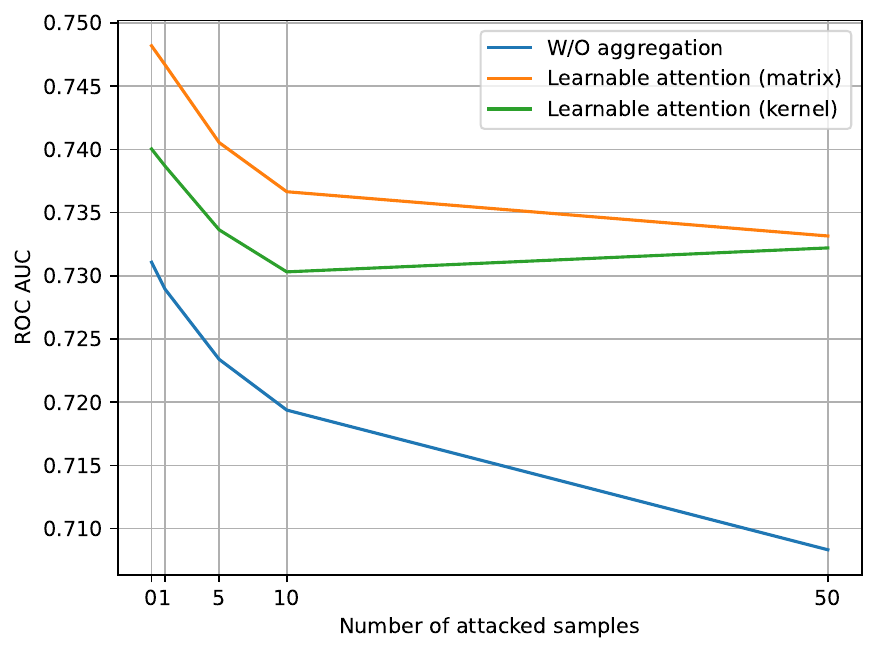}
\captionsetup{font=small}
\caption{Decreasing ROC-AUC metric under random attack with different number of attacked samples for CoLES model, Churn dataset, global validation pipeline. The number of attacked samples 0 is equal to the case without an attack.}
\label{fig:attack_rand}
\end{figure}

The second way to validate the robustness of the external context models is to perform an adversarial attack on categorical features by randomly changing their values. The metrics decrease for these adversarial attacks are presented in Figures~\ref{fig:attack_rand}. In this experiment, the metrics dropped less dramatically than in the previous one because we didn't generate specially perturbed transactions, but instead used random ones. The experiment setup is close to the previous one. Learnable attack quality decreases much less than in the vanilla model as the number of perturbed transactions increases.


\paragraph{External information with lack of relevant events}

\begin{figure}[!t]
\centering
\includegraphics[width=.9\linewidth]{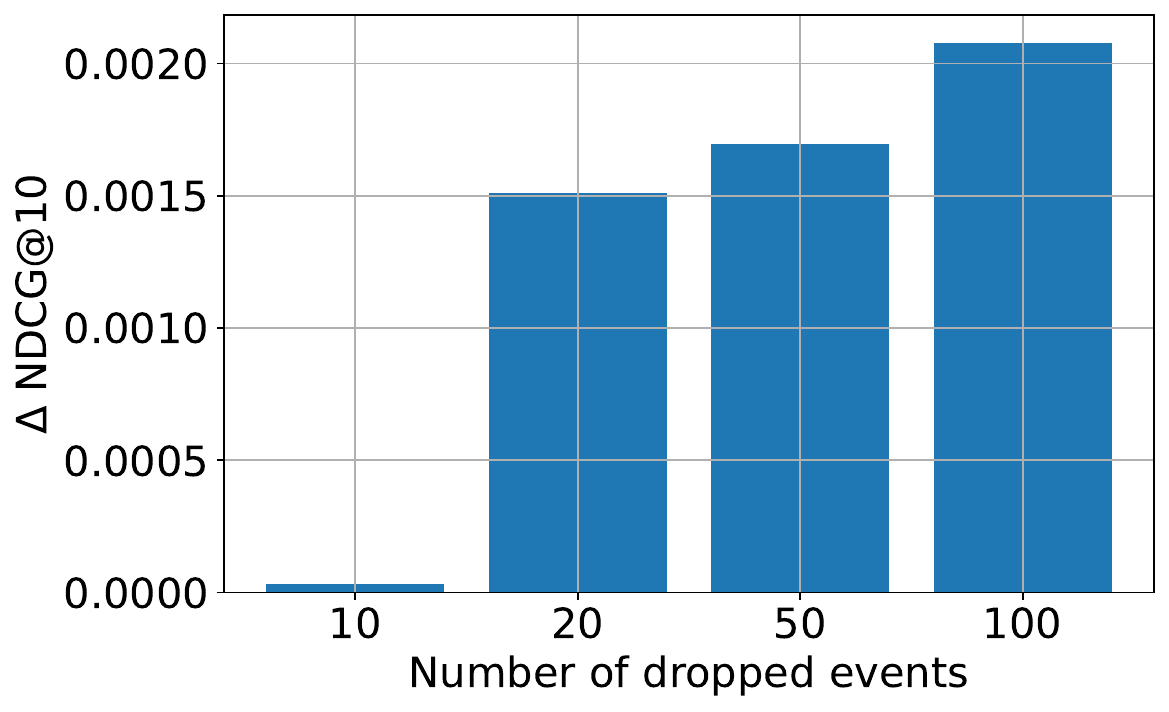}
\captionsetup{font=small}
\caption{Improving NDCG@10 for the RecSys pipeline and the MovieLens 1M dataset for Learnable attention aggregation. We dropped the last \textit{Number of dropped events} of every sequence to assess the external aggregation for cases without relevant events. As we can see, the external context improves the model's performance on these sequences.}
\label{fig:rs_dropped}
\end{figure}

One of the aims of external context aggregation is to improve model predictions for users with long lags between the last event and the moment of the next prediction. Aggregating other users' external context with relevant events occurring in the absence of the target sequence can increase the model quality. We conduct experiments by dropping several recent events before the predicted event using local validation. As shown in Figure~\ref{fig:rs_dropped}, we observe improvements in model quality across different numbers of dropped events. The external context not only improves predictions but also does so more effectively with more dropped events.

\subsubsection{Computational overhead for the external aggregation}
\label{sec: res_time}

During inference, incorporating external context incurs only minor computational overhead if the overall procedure is implemented properly.
However, during training, additional expenses can outweigh potential benefits.
Below, we numerically estimate the computational overhead introduced by external aggregation.

The full procedure consists of the following training and inference stages:
\begin{itemize}
    \item \textit{Pretrain SSL} is the training encoder with self-supervised loss. In this stage, we use a pre-trained encoder to obtain only internal representation vectors.
    \item \textit{Pretrain Pool SSL} is training pooling learnable parameters (only for learnable pooling methods) with self-supervised loss. In this stage, we obtain a pre-trained encoder with pooling layers to produce both internal and external representation vectors.
    \item \textit{Fine-tune, frozen} and \textit{Fine-tune, unfrozen} are training the head over frozen or unfrozen backbone with pooling on the next event prediction task, respectively. At this stage, we train the encoder's head for the local validation task.
    \item \textit{Vanilla inference} is the vanilla generative baseline-encoder embeddings, an internal representation vector.
    \item \textit{Inference with aggregations} is the generation of both internal and external embeddings, proposed in this article.
\end{itemize}
The first stage, \textit{Pretrain  SSL}, is simply the training of a basic model; the others are additional costs that may be required to use external aggregation approaches.

\input{tables/time}

The training and inference times are shown in Table~\ref{tab:time} for the training and test sets of the Churn dataset, respectively.
We see that all training times are comparable, but additional training with unfreezing requires significant additional resources while simultaneously boosting model quality.

%% file: tables/dataset.tex
\begin{table*}[!h]
\captionsetup{font=small}
\caption{Basic statistics for event sequences datasets}
\label{tab:datasets}
\centering
\resizebox{\textwidth}{!}{%
\begin{tabular}{lccccccccc}
\hline
~ & Churn & Default & HSBC  & Taobao & ML-1M & Beauty & Beer Advocate & 30Music & MovieLens-20M  $ $\\
\hline
Pipeline & Class. &  Class. & Class. &  Class. &  Both & RecSys & RecSys & RecSys & RecSys \\
N. of events & 490K & 2M & 234K  & 7.9M & 1M & 6.6M & 1.4M & 31M & 20M \\
Min. seq. length & $1$ & $300$ & $1$  & $1$ & $20$ & $5$ & $20$ & $20$ & $20$  \\
Max. seq. length & $784$ & $300$ & $3467$  & $21K$ & $2314$ & $2351$ & $4768$ & $89K$ & $9254$  \\
Med. seq. length & $83$ & $300$ & $40$ & $473$ & $96$ & $7$ & $72$ & $400$ & $68$   \\
Class imbalance & $35:65$ & $ 4:96$ & $8\!:\!92$ & $75\!:\!25$ & $75\!:\!25$ & -  & - & - & -  \\
\hline
\end{tabular}}
\end{table*}

%% file: tables/hyperparams.tex
\begin{table}[!ht]
\caption{Summary of hyperparameters}
\label{tab:hyperparams_trans}
\centering
\resizebox{\columnwidth}{!}{%
\begin{tabular}{lllccc}
    \hline
    Pipeline & Architecture & Dataset(s) & Hidden & Batch & Number of
    \\
    &&& dimension & size & training epochs \\
    \hline
    \noalign{\vskip 3pt}
    Classification & LSTM & Churn, HSBC, &1024 & 128 & 60 \\
     &  &  Taobao &  &  &  \\
     &  &  MovieLens &  &  &  \\
    & GRU & Default & 800 & 128 & 60 \\
    & Transformer & Churn, Default, & 512 & 128 & 60 \\
    &  &  HSBC, Taobao &  &  &  \\
    &  &  MovieLens 1M &  &  &  \\
    \hline
    RecSys & SASRec & MovieLens 1M & 128 & 32 & 150 \\
    &  & Beauty & 128 & 32 & 20 \\
    &  & Beer Advocate & 128 & 32 & 50 \\
    &  & 30 Music & 128 & 32 & 20 \\
    &  & MovieLens 20M & 128 & 32 & 20 \\
    \hline
\end{tabular}}
\end{table}


%% file: tables/ranks_4.tex
\begin{table*}[!t]
\captionsetup{font=small}
\caption{\selectfont Summarized metrics ($\uparrow$) with ranks in parentheses ($\downarrow$) based on the ROC-AUC metric for all the datasets for the classification pipeline. All metrics have been rounded to four decimal places before getting ranks. 
The best values are \textbf{highlighted}, the second-best values are \underline{underlined}.}
\label{tab:ranks}
\centering
\resizebox{\textwidth}{!}
{
\begin{tabular}{lccccccccccc}
\toprule
\multicolumn{1}{c}{\textbf{Aggregation}} &
\multicolumn{2}{c}{\textbf{Churn}} &
\multicolumn{2}{c}{\textbf{Default}} &
\multicolumn{2}{c}{\textbf{HSBC}} &
\multicolumn{2}{c}{\textbf{Taobao}} &
\multicolumn{2}{c}{\textbf{MovieLens1M}} &
\multicolumn{1}{c}{\textbf{Mean}} \\
\cmidrule(lr){2-3} \cmidrule(lr){4-5} \cmidrule(lr){6-7} \cmidrule(lr){8-9} \cmidrule(lr){10-11} 
&
\textbf{Global} & \textbf{Local} &
\textbf{Global} & \textbf{Local} &
\textbf{Global} & \textbf{Local} &
\textbf{Global} & \textbf{Local} &
\textbf{Global} & \textbf{Local} & \textbf{rank} \\
\midrule
Without context & 0.7432 & 0.7004 & 0.5493 & 0.7526 & 0.6930 & 0.8980 & 0.7137 & 0.7996 & 0.7935 & 0.6899 & 6.7 \\
Mean & 0.7728 & 0.7102 & 0.5557 & 0.7526 & 0.7335 & 0.9007 & 0.7138 & 0.8014 & 0.7948 & \second{0.6956} & 3.8 \\
Max & \second{0.7737} & \second{0.7113} & 0.5413 & 0.7524 & 0.7234 & 0.9015 & 0.7138 & 0.8013 & 0.7953 & \best{0.6966} & 4.0 \\
Exp Hawkes & 0.7701	& 0.7102 & 0.5577 & 0.7523 & 0.7343 & \best{0.9207} & 0.7138 & 0.8014 & 0.7948 & 0.6956 & 3.5 \\
Attention Hawkes & 0.7609 & 0.7098 & 0.5556 & 0.7523 & 0.6706 & 0.9184 & 0.7099 & \best{0.8023} & 0.7957 & 0.6940 & 5.0 \\
Attention & 0.7598 & 0.7095 & \best{0.5629} & 0.7487 & 0.6687 & 0.8996 & 0.7099 & \best{0.8023} & 0.7889 & 0.6940 & 5.9 \\
Learnable exp Hawkes* & 0.7670 & \best{0.7191} & 0.5582 & \best{0.7586} & \best{0.7493} & \second{0.9189} & \best{0.7181} & 0.7846 & \second{0.7997} & 0.6758 & \second{3.2} \\
Learnable attention* & \best{0.7768} & 0.7102 & \second{0.5594} & \second{0.7527} & \second{0.7471} & 0.9009 & \second{0.7171} & \second{0.8015} & \best{0.8019} & \second{0.6956} & \best{2.3} \\
\bottomrule
\end{tabular}
}
\end{table*}

%% file: tables/prauc.tex
\begin{table}[!ht]
\centering
\captionsetup{font=small}
\caption{PR-AUC metrics for classification pipeline under global validation, trained with contrastive loss and RNN encoder. The best values are \textbf{highlighted}, and the second-best values are \underline{underlined}.}
\label{tab:pr-auc}
\begin{tabular}{lccc}
\toprule
Method & Churn & Default & HSBC \\
\midrule
Without context & 0.7922 & 0.0580 & 0.1540 \\
Mean & \underline{0.8277} & 0.0564 & \textbf{0.2368} \\
Max & 0.8184 & 0.0494 & 0.2058 \\
Exp Hawkes & 0.8140 & \underline{0.0708} & 0.1389 \\
Attention Hawkes & 0.7963 & 0.0625 & 0.3310 \\
Attention & 0.8085 & 0.0503 & 0.1472 \\
Learnable exp Hawkes* & 0.8120 & 0.0613 & 0.1282 \\
Learnable attention* & \textbf{0.8303} & \textbf{0.0727} & \underline{0.2248} \\
\bottomrule
\end{tabular}
\end{table}

%% file: tables/external_baselines.tex
\begin{table}[h!]
\centering
\caption{Accuracy score for the most popular transaction baseline experiment.}
\label{tab: baseline}
\begin{tabular}{lccc}
\hline
Dataset & Churn & Default & HSBC \\
\hline
internal & 0.194 & 0.185 & 0.650 \\
internal + external & 0.282 & 0.356 & 0.636 \\
\hline
\end{tabular}
\end{table}

%% file: tables/only_ext.tex
\begin{table}[!h]
\centering
\captionsetup{font=small}
\caption{Comparing predictions built only from external aggregation embeddings for RecSys (MovieLens 1M dataset) and the Sequential (Churn dataset) pipeline.}
\label{tab:only_external}
\begin{tabular}{lccc}
\hline
Pipeline & RecSys & \multicolumn{2}{c}{Sequential} \\
Validation &  & Global valid. & Local valid. \\
Agg. method & NDCG@10 & ROC-AUC & ROC-AUC \\
\hline
\multicolumn{4}{c}{Only internal context}\\
\hline
-                 & 0.0775 & 0.7432 & 0.7004 \\
\hline
\multicolumn{4}{c}{Only external context}\\
\hline
Mean                 & 0.0092 & 0.5004 & 0.6503 \\
Attention            & 0.0546 & 0.5407 & 0.7097 \\
Learnable attention  & 0.0561 & 0.5279 & 0.6918 \\

\hline
\multicolumn{4}{c}{Both internal and external context}\\
\hline
Mean                 & 0.0775 & 0.7728 & 0.7102 \\
Attention            & 0.0776 & 0.7598 & 0.7095 \\
Learnable attention  & 0.0778 & 0.7768 & 0.7102 \\

\hline
\end{tabular}
\end{table}

%% file: tables/ar.tex
\begin{table}[!ht]
\captionsetup{font=small}
\caption{Performance metrics for classification pipeline with generative loss across datasets. The best values are \textbf{highlighted}, the second-best values are \underline{underlined}.}
\label{tab:generative_metrics}
\centering
\resizebox{\columnwidth}{!}
{
\begin{tabular}{lcccccc}
\toprule
\multicolumn{1}{c}{\textbf{Aggregation}} &
\multicolumn{2}{c}{\textbf{Churn}} &
\multicolumn{2}{c}{\textbf{Default}} &
\multicolumn{2}{c}{\textbf{HSBC}} \\
\cmidrule(lr){2-3} \cmidrule(lr){4-5} \cmidrule(lr){6-7}
&
\textbf{Global} & \textbf{Local} &
\textbf{Global} & \textbf{Local} &
\textbf{Global} & \textbf{Local} \\
\midrule
\multicolumn{7}{c}{\textbf{Baseline}} \\
\midrule
Without context & 0.7148 & 0.7209 & 0.5153 & 0.7589 & 0.6405 & 0.9204 \\
\midrule
\multicolumn{7}{c}{\textbf{Without fine-tuning}} \\
\midrule
Mean & 0.7613 & 0.7226 & 0.5153 & 0.7588 & \best{0.7064} & 0.9213 \\
Max & \best{0.7655} & 0.7217 & 0.5207 & 0.7587 & 0.6872 & \best{0.9216} \\
Exp Hawkes & 0.7631 & 0.7119 & \second{0.5319} & \second{0.7596} & 0.6905 & \second{0.9214}\\
Attention Hawkes & 0.7212 & 0.6687 & 0.5234 & 0.7562 & 0.6826 & 0.9165	 \\
Attention & 0.7246 & 0.7177 & 0.5246 & 0.7552 & 0.6694 & 0.9163 \\
Learnable exp Hawkes*  & 0.7417 & 0.6853 & 0.5051 & 0.7189 & \second{0.6996} & 0.9121 \\
Learnable attention* & 0.7533 & 0.7226 & 0.5250 & 0.7575 & 0.6981 & 0.9210 \\
\midrule
\multicolumn{7}{c}{\textbf{With fine-tuning}} \\
\midrule
Mean & 0.7559 & 0.7160 & 0.5153 & 0.7551 & \best{0.7064} & 0.9178 \\
Max & \second{0.7647} & 0.7152 & 0.5207 & 0.7545 & 0.6872 & 0.9183 \\
Exp Hawkes & 0.7628 & \best{0.7231} & \second{0.5319} & \second{0.7596} & 0.6906 & 0.9223 \\
Attention Hawkes & 0.7210 & 0.6733 & 0.5269 & 0.7545 & 0.6826 & 0.9158 \\
Attention & 0.7246 & 0.7151 & 0.5246 & 0.7521 & 0.6694 & 0.9160 \\
Learnable exp Hawkes*  & 0.7453 & 0.6920 & 0.5051 & 0.7188 & 0.6797  & 0.8959 \\
Learnable attention* & 0.7527 & \second{0.7230} & \best{0.5430} & \best{0.7606} & 0.6658 & 0.9184 \\
\bottomrule
\end{tabular}
}
\end{table}

%% file: tables/transfromers_2.tex
\begin{table}[!h]
\captionsetup{font=small}
\caption{\selectfont ROC-AUC values $(\uparrow)$ for global and local embedding validation results obtained with the \textbf{Transformer} encoder. 
The results are averaged over three runs. 
The best values are \textbf{highlighted}, and the second-best values are \underline{underlined}.}
\label{tab:metrics_table_transformer}
\centering
\resizebox{\columnwidth}{!}{%
\setlength{\tabcolsep}{3pt}
\begin{tabular}{lccccc}
\hline
Dataset & Churn & Default & HSBC & Taobao & ML1M \\
\hline
\multicolumn{6}{c}{\textit{Global validation}} \\
\hline
Without context & 0.6803 & 0.5379 & 0.6548 & 0.7486 & 0.7505 \\
Mean & \textbf{0.7308} & 0.5414 & \underline{0.7281}  & \textbf{0.7742} & 0.7833 \\
Max & 0.7148 & 0.5461 & 0.7062 & \textbf{0.7742} & 0.7834 \\
Exp Hawkes & \underline{0.7253}  & 0.5223  & 0.5660  & \textbf{0.7742}  & 0.7829 \\
Attention Hawkes & 0.6928  & 0.4872  & 0.5532  & \underline{0.7714}  & 0.7843 \\
Attention & 0.6758 & \textbf{0.5580} & 0.6733 & \underline{0.7714} & 0.7825 \\
Learnable exp Hawkes* & 0.7144  & 0.5187  & 0.6130  & 0.7696  & \textbf{0.7889} \\
Learnable attention* & 0.7193 & \underline{0.5523} & \textbf{0.7462} & 0.7668 & \underline{0.7840} \\

\hline
\multicolumn{6}{c}{\textit{Local validation}} \\
\hline
Without context & 0.6865 & 0.7024 & 0.9121 & 0.7446 & 0.6801 \\
Mean & \underline{0.7249} & \underline{0.7539} & 0.9145 & 0.7818 & 0.6761 \\
Max & 0.7242 & 0.7533 & \textbf{0.9164} & 0.7854 & 0.6654 \\
Exp Hawkes & 0.7156  & 0.7361  & 0.9148  & 0.7742  & \textbf{0.6935}  \\
Attention Hawkes & 0.7066  & 0.7314 & 0.9110 & 0.7714  & \underline{0.6922}  \\
Attention & 0.7179 & 0.7493 & 0.9153 & 0.7776 & 0.6672 \\
Learnable exp Hawkes* & 0.6830  & 0.6983  & 0.9089  & \textbf{0.7969}  & 0.6712  \\
Learnable attention* & \textbf{0.7254} & \textbf{0.7552} & \underline{0.9155} &  \underline{0.7873} & 0.6816 \\
\hline
\end{tabular}%
}
\end{table}

%% file: tables/time.tex
\begin{table}[t]
\captionsetup{font=small}
\caption{\selectfont Time of different training and inference stages for the Churn dataset}
\label{tab:time}
\centering
\resizebox{\columnwidth}{!}{%
\begin{tabular}{lccc}
\hline
Stage & Mean & Exponential & Learnable  \\
 &  & Hawkes & attention \\
\hline
\emph{Training, minutes} &&& \\
Pretrain SSL & 8.295 & 8.295 & 8.295 \\
Pretrain Pool SSL & - & - & 7.541  \\
Fine-tune, freeze & 7.276 & 9.326 & 11.524 \\
Fine-tune, unfreeze & 15.04 & 19.874 & 20.286  \\
\hline
\emph{Inference, seconds} &&& \\
Vanilla inference & 0.878 & 0.878 & 0.878  \\
Inference with aggregations & 0.917 & 0.922 & 0.926  \\
\hline
\end{tabular}}
\end{table}

%% file: chapters_sigmod/05_conclusion.tex
\section{Conclusion and discussions}

We consider the problem of creating representations for non-uniform-in-time event sequences from various domains. 
Our model enhances the quality of existing approaches by introducing an aggregation procedure for external contextual information, which represents global context, such as macroeconomic parameters~\cite{begicheva2021bank} and the behavior of similar users. 
The paper considers a range of methods for such aggregations based on pooling, attention mechanisms, and those inspired by the self-exciting temporal point process. 

The representations obtained using the addition of external context vectors improve model performance on downstream problems, particularly those in the bank transaction data domain and in e-commerce.
The most effective results were achieved with a Learnable attention method, which identifies the clients closest to the target user.
Further improvement can be achieved by fine-tuning the model to suit the aggregation pipeline better.


Beyond the primary comparison of aggregation strategies, our comprehensive evaluations consistently validate the practical utility of external context. Sensitivity analyses across different backbones (RNN vs. Transformer), pretraining objectives (contrastive vs. generative), and domain-specific pipelines (Classification and RecSys) confirm that Learnable attention is statistically significantly the top-rated method, demonstrating advanced performance for the majority of considered examples. Additional robustness checks further demonstrate that external aggregation reduces performance degradation under adversarial perturbations, random feature noise, and distribution shifts.

From both theoretical and empirical perspectives, the optimal number of external sequences follows a clear pattern of diminishing returns. The Gaussian process framework introduced in Section~\ref{sec:theory} formalizes this behavior: external series act as co-kriging observations that capture shared macro-trends, and the integrated extrapolation risk converges exponentially to its lower bound as the number of external series increases. Theoretical analysis indicates that a moderate range of $K \in [10, 50]$ is typically sufficient to achieve near-optimal risk reduction, provided a non-negligible shared trend exists. This insight aligns closely with our empirical sensitivity study, where marginal gains in downstream metrics flatten once the context size reaches approximately $n=100\text{--}500$ sequences. Consequently, in production settings, aggregating over a bounded subset of recent users strikes a principled balance between representational gain and computational overhead.


%% file: chapters_sigmod/99_appendix.tex
\section{Additional results}

\subsection{Significance}
\label{sec:cd}

In this paragraph, we present Table~\ref{tab:cd} of the external aggregation method pairwise comparison. The table was compiled based on all RecSys and Classification pipeline experiments, including transformer architecture and generative loss experiments. The results of the table are equal and complement Figure~\ref{fig:cd}.

\input{tables/cd}

\subsection{Classification pipeline additional results}
\label{sec:add_results}

In this paragraph, we present the full experimental Table~\ref{tab:metrics} for the classification pipeline with contrastive loss, which was run with three different random seeds. 

\input{tables/metric_table_v3}

\subsection{Derivation of the BLUP risk representation}
\label{sec:theory}

$X_D$ denotes the vector of all observations from the training sample $D$.
For a fixed prediction point $t$, consider the target random variable $x_0(t)$ and the observation vector $X_D$.
For the Gaussian model under consideration, the random vector
\begin{equation}
\begin{pmatrix}
x_0(t)\\
X_D
\end{pmatrix}
\end{equation}
is jointly Gaussian with zero mean and covariance matrix
\begin{equation}
\begin{pmatrix}
v(t) & c_D(t)^\top\\
c_D(t) & \Sigma_D
\end{pmatrix},
\end{equation}
where
\begin{equation}
v(t)=\operatorname{Var}(x_0(t)),
\end{equation}
\begin{equation}
c_D(t)=\operatorname{Cov}(x_0(t),X_D),
\end{equation}
and
\begin{equation}
\Sigma_D=\operatorname{Cov}(X_D,X_D).
\end{equation}
By the standard conditioning formula for jointly Gaussian random variables, the BLUP is given by
\begin{equation}
\hat{x}_0(t)=\mathbb{E}[x_0(t)\mid X_D]=c_D(t)^\top\Sigma_D^{-1}X_D.
\end{equation}
The corresponding posterior variance is
\begin{equation}
\operatorname{Var}(x_0(t)\mid X_D)=v(t)-c_D(t)^\top\Sigma_D^{-1}c_D(t).
\end{equation}

Since $\hat{x}_0(t)=\mathbb{E}[x_0(t)\mid X_D]$, the conditional mean squared error is equal to the posterior variance:
\begin{equation}
\mathbb{E}\left[(\hat{x}_0(t)-x_0(t))^2 \mid X_D\right]
=
\operatorname{Var}(x_0(t)\mid X_D).
\end{equation}
Therefore, for a fixed training sample $D$, we obtain
\begin{equation}
R(D,h,\delta)
=
\int_h^{h+\delta}\operatorname{Var}(x_0(t)\mid X_D)\,dt
\end{equation}

Thus, minimizing the integrated extrapolation risk is the same as minimizing the posterior variance of the target process over the prediction interval.
This explains why external observations are useful: they provide additional information about the shared latent component and therefore reduce the uncertainty of the prediction.

\subsection{Interpretability of external aggregation}
\label{sec: interpretability}

We consider three variants for the interpretability: (1) how external aggregation helps to increase robustness to domain shifts, (2) how the learned attention matrices for the external aggregations look, (3) how useful the external features are for the gradient boosting head compared to the internal values.

\paragraph{Effect of external aggregation under distribution shift}

To evaluate the usefulness of the external context in cases of distribution drift, we conduct an experiment that measures model quality after overall shifts in the data distribution. 
The shifts are identified as change points in the multivariate series of mean embeddings from all selected users, detected using PELT~\cite{killick2012optimal}.
This robust method for change-point detection delivers competitive performance on multivariate time series and requires only a small number of hyperparameters to tune~\cite {van2020evaluation}.
We followed standard recommendations with a penalty of $70$ and a minimum interval size of $20$. 
The detected change points should indicate external events affecting the overall user population. We estimate ROC-AUC metrics for the baseline model and for Mean and Learnable attention external aggregations using a local validation procedure before and after detected change points. 

\begin{figure}[!t]
     \centering
     \includegraphics[width=0.9\columnwidth]{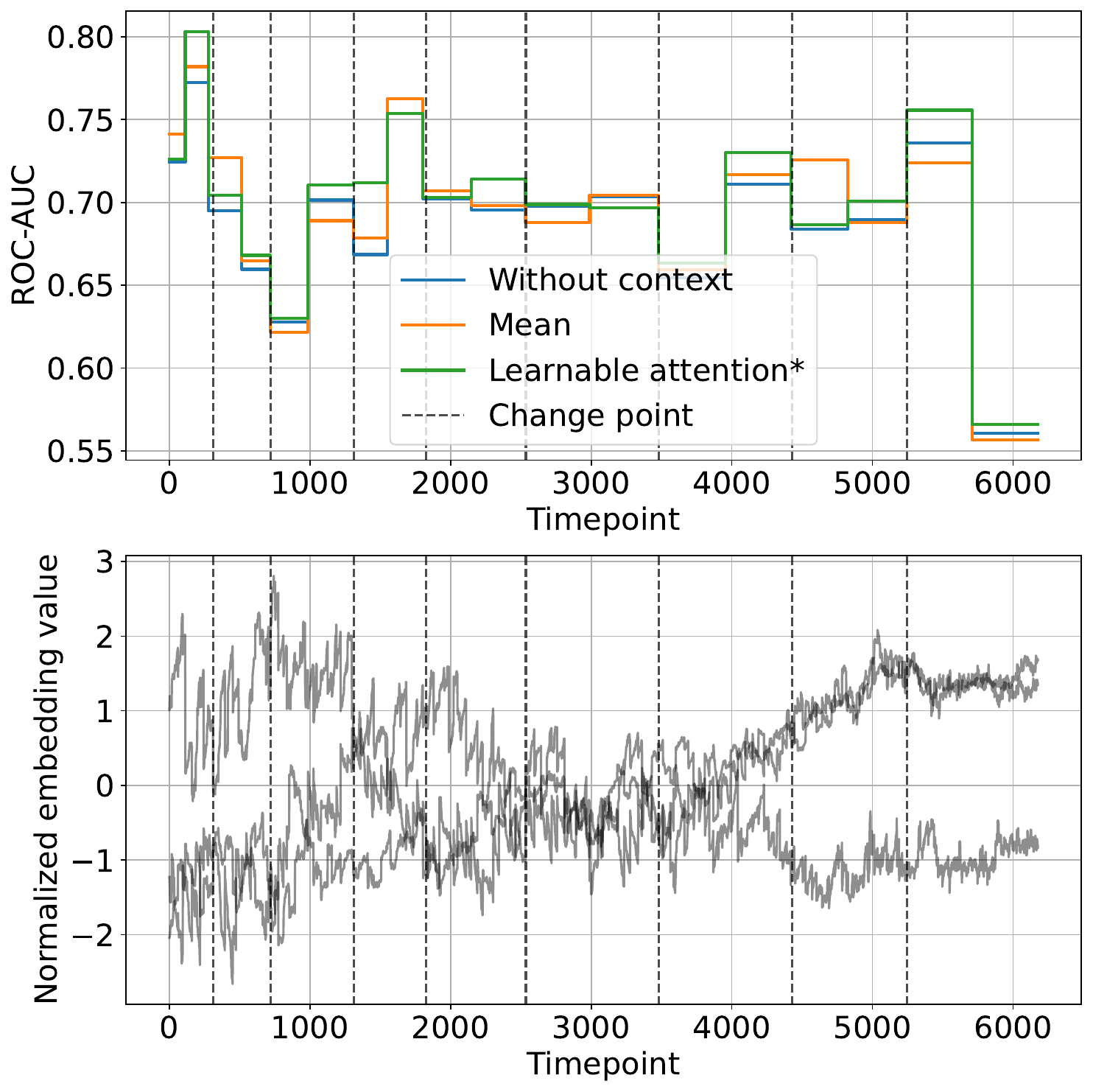}
     \caption{Top plot: local validation ROC-AUC changes after external events, detected with change points, Mean and Learnable attention aggregation and modeling without aggregation. Bottom plot: five randomly selected normalized axes of time embeddings, used for change point detection.}
     \label{fig:cpd}
\end{figure}

The results for the Churn dataset are presented in Figure~\ref{fig:cpd}. 
After change points, the quality of a model without aggregation typically drops. 
External aggregations mostly mitigate this quality loss, making the model more robust to distribution shifts. 
This suggests that external context vectors serve as stable priors, helping the model retain predictive power even as the population behavior evolves.

\paragraph{Structure of user-relation attention matrices}

\begin{figure}[!b]
\centering
  \centering
  \includegraphics[width=\linewidth]{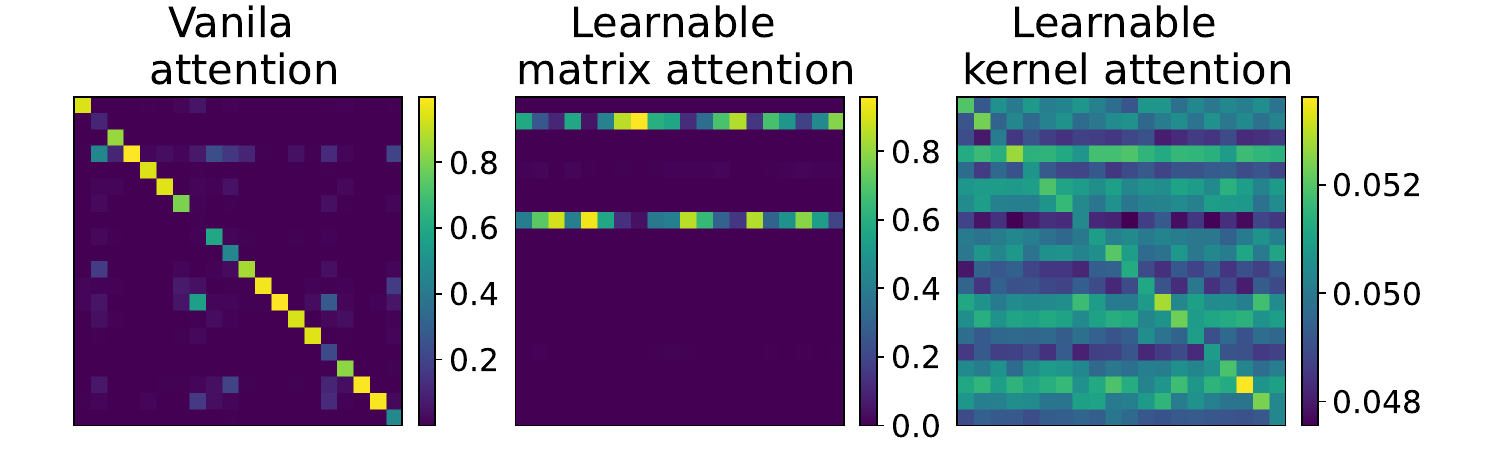}
  \caption{Attention matrices obtained in different methods. From left to right: Vanilla attention, Learnable Matrix attention, and Learnable Kernel attention}
  \label{fig:attn_matrices}
\end{figure}

To examine uncovered dependencies among users, we analyzed the cross-user attention matrices obtained. 
For a random time point in the Churn dataset and a random subsample of users, the analysis considers $\mathrm{softmax}(H_t^T H_t)$ for the \emph{Attention}; $\mathrm{softmax}(H_t^T A H_t)$ for \emph{Matrix Attention}, and $\mathrm{softmax}(\langle \phi(H_t), \phi(H_t) \rangle)$ for \emph{Kernel Attention}. These matrices, which reveal how different users are weighted in relation to one another, are depicted in Figure \ref{fig:attn_matrices}.

In the \emph{Attention} case, the outcome is expected: each user is most similar to themselves, as dictated by the contrastive learning objective. The \emph{Matrix Attention} method tends to emphasize certain users in the sample who are deemed most relevant to all others, leading to similarity collapse. In contrast, the \emph{Kernel Attention} method produces a more uniform distribution of user similarities with a tendency for the major diagonal dominance.

These findings highlight a critical insight: constraining embeddings to be maximally self-similar strangles relational expressivity. By contrast, Learnable Kernel Attention leverages a complete geometry of the user manifold, transforming a sparse diagonal view into a dense, information-rich affinity graph. The resulting representations not only preserve nuanced inter-user structure but also yield consistent downstream gains, as shown above.

\paragraph{External feature importance}
\label{sec:add_res}

\input{tables/shap}

To assess the importance of added vectors, we analyzed SHAP interaction values~\cite{shap2017}. 
These values were computed for a gradient boosting classifier in a global validation task on the Churn dataset. 
We consider two sets of features: \emph{INT} corresponds to the internal representation features, and \emph{EXT} corresponds to the external information features after aggregation.
\emph{ABS} is the sum of absolute SHAP values for corresponding features (ABS INT is for internal and ABS EXT is for external features).
\emph{Interaction} is the sum of corresponding submatrices in SHAP interaction values with INT vs INT corresponding to internal information, EXT vs EXT --- to external information, and INT vs EXT --- to interactions between internal and external information.

The results are presented in Figure~\ref{tab:shap}.
Absolute SHAP values show that the external information vector contributes to the final result, while being less important with smaller values compared to internal information. 
The same result is also evident from the summation of SHAP interaction values across external vector features. 
Also, it is interesting to see that there is little correspondence between internal and external features, highlighting that they capture different aspects of the behavior dynamics.

%% file: tables/cd.tex
\begin{table}[!h]

\centering
\caption{Pairwise statistically significant comparison of Wilcoxon-Holm analysis, $\alpha=0.05$}
\label{tab:cd}
\begin{tabular}{llcc}
\toprule
{Method A} & {Method B} & p-value & Significant \\
\midrule
Learnable attention* & Without context & 1.770e-08 & True \\
Attention & Learnable attention* & 1.954e-04 & True \\
Mean & Without context & 3.221e-04 & True \\
Max & Without context & 6.440e-04 & True \\
Attention & Mean & 3.597e-03 & True \\
Attention & Max & 5.442e-03 & True \\
Learnable attention* & Mean & 5.885e-03 & True \\
Learnable attention* & Max & 7.917e-03 & True \\
Attention & Without context & 2.744e-02 & False \\
Max & Mean & 5.563e-01 & False \\
\bottomrule
\end{tabular}
\end{table}

%% file: tables/metric_table_v3.tex
\begin{table*}[!h]
\caption{\selectfont ROC-AUC values $(\uparrow)$ for global and local embedding validation results obtained with \textbf{RNN encoder}. 
The results are averaged by three runs and are given in the format $mean \pm std$. 
The best values are \textbf{highlighted}, and the second-best values are \underline{underlined}.}
\label{tab:metrics}
\centering
\begin{tabular}{lccccc}
\hline
Dataset & Churn & Default & HSBC & Taobao & MovieLens1M\\
\hline
\multicolumn{6}{c}{\textit{Global validation}} \\
\hline
Without context & 0.7432 $\pm$ 0.009 & 0.5493 $\pm$ 0.026 & 0.6930 $\pm$ 0.041 & 0.7137 $\pm$ 0.016 & 0.7935 $\pm$ 0.011 \\
Mean & 0.7728 $\pm$ 0.004 & 0.5557 $\pm$ 0.004 & 0.7335 $\pm$ 0.017 & 0.7138 $\pm$ 0.016 & 0.7948 $\pm$ 0.009 \\
Max & \underline{0.7737 $\pm$ 0.021} & 0.5413 $\pm$ 0.018 & 0.7234 $\pm$ 0.020 & 0.7138 $\pm$ 0.016 & 0.7953 $\pm$ 0.010 \\
Exp Hawkes & 0.7701 $\pm$ 0.003 & 0.5577 $\pm$ 0.029 & 0.7343 $\pm$ 0.031 & 0.7138 $\pm$ 0.016 & 0.7948 $\pm$ 0.009 \\
Attention Hawkes & 0.7609 $\pm$ 0.007 & 0.5556 $\pm$ 0.022 & 0.6706 $\pm$ 0.022 & 0.7099 $\pm$ 0.014 & 0.7957 $\pm$ 0.012 \\
Attention & 0.7598 $\pm$ 0.014 & \textbf{0.5629 $\pm$ 0.008} & 0.6687 $\pm$ 0.013 & 0.7099 $\pm$ 0.014 & 0.7889 $\pm$ 0.006 \\
Learnable exp Hawkes* & 0.7670 $\pm$ 0.005 & 0.5582 $\pm$ 0.024 & \textbf{0.7493 $\pm$ 0.019} & \textbf{0.7181 $\pm$ 0.012} & \underline{0.7997 $\pm$ 0.009} \\
Learnable attention* & \textbf{0.7768 $\pm$ 0.013} & \underline{0.5594 $\pm$ 0.003} & \underline{0.7471 $\pm$ 0.018} & \underline{0.7171 $\pm$ 0.013} & \textbf{0.8019 $\pm$ 0.011} \\
\hline
\multicolumn{6}{c}{\textit{Local validation}} \\
\hline
Without context & 0.7004 $\pm$ 0.002 & 0.7526 $\pm$ 0.002 & 0.8980 $\pm$ 0.009 & 0.7996 $\pm$ 0.002 & 0.6899 $\pm$ 0.002 \\
Mean & 0.7102 $\pm$ 0.003 & 0.7526 $\pm$ 0.001 & 0.9007 $\pm$ 0.003 & 0.8014 $\pm$ 0.002 & \underline{0.6956 $\pm$ 0.002} \\
Max & \underline{0.7113 $\pm$ 0.003} & 0.7524 $\pm$ 0.001 & \underline{0.9015 $\pm$ 0.003} & 0.8013 $\pm$ 0.002 & \textbf{0.6966 $\pm$ 0.002} \\
Exp Hawkes & 0.7102 $\pm$ 0.002 & 0.7523 $\pm$ 0.001 & 0.9207 $\pm$ 0.003 & 0.8014 $\pm$ 0.002 & 0.6956 $\pm$ 0.002 \\
Attention Hawkes & 0.7098 $\pm$ 0.003 & 0.7523 $\pm$ 0.001 & 0.9184 $\pm$ 0.003 & \textbf{0.8023 $\pm$ 0.002} & 0.6940 $\pm$ 0.002 \\
Attention & 0.7095 $\pm$ 0.003 & 0.7487 $\pm$ 0.001 & 0.8996 $\pm$ 0.001 & \textbf{0.8023 $\pm$ 0.002} & 0.6940 $\pm$ 0.002 \\
Learnable exp Hawkes* & \textbf{0.7191 $\pm$ 0.002} & \textbf{0.7586 $\pm$ 0.001} & \textbf{0.9189 $\pm$ 0.003} & 0.7846 $\pm$ 0.005 & 0.6758 $\pm$ 0.001 \\
Learnable attention* & 0.7102 $\pm$ 0.002 & \underline{0.7527 $\pm$ 0.001} & 0.9009 $\pm$ 0.004 & \underline{0.8015 $\pm$ 0.002} & \underline{0.6956 $\pm$ 0.002} \\
\hline
\end{tabular}
\end{table*}

%% file: tables/shap.tex

\begin{figure}[!t]
  \centering

  \begin{subfigure}[t]{0.45\textwidth}
    \centering
    \resizebox{\columnwidth}{!}{
    \begin{tabular}{lccc}
    \toprule
    Interactions & Mean & Max & Learn. attention \\
    \midrule
    ABS INT       & 0.738 $\pm$ 0.009 & 0.804 $\pm$ 0.028 & 0.766 $\pm$ 0.004 \\
    ABS EXT       & 0.276 $\pm$ 0.004 & 0.177 $\pm$ 0.007 & 0.250 $\pm$ 0.027 \\
    INT vs INT    & 0.087 $\pm$ 0.013 & 0.080 $\pm$ 0.013 & 0.069 $\pm$ 0.016 \\
    EXT vs EXT    & 0.038 $\pm$ 0.018 & 0.038 $\pm$ 0.023 & 0.024 $\pm$ 0.030 \\
    INT vs EXT    & -0.007 $\pm$ 0.005 & -0.002 $\pm$ 0.002 & 0.003 $\pm$ 0.004 \\
    \bottomrule
    \end{tabular}}
    \label{tab:shift_auc}
  \end{subfigure}
  \\
  \begin{subfigure}[t]{0.3\textwidth}
    \centering
    \includegraphics[width=\linewidth]{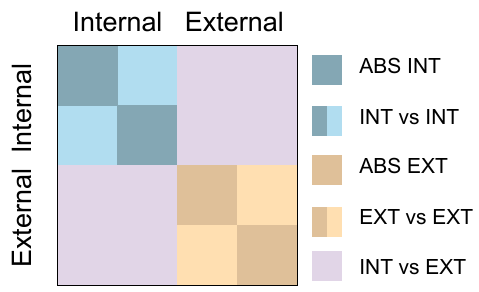}
    \label{fig:performance_plot}
  \end{subfigure}
  \captionsetup{font=small}
  \caption{SHAP values for different methods in the global validation task on the Churn dataset, average over three runs. The columns refer to specific aggregation methods. Top: obtained values, bottom: aggregation scheme.}
  \label{tab:shap}
\end{figure}